\documentclass[letterpaper, 10 pt, conference]{ieeeconf}  % Comment this line out if you need a4paper

\IEEEoverridecommandlockouts                              % This command is only needed if 
\usepackage{graphics} % for pdf, bitmapped graphics files
\usepackage{epsfig} % for postscript graphics files
\usepackage{times} % assumes new font selection scheme installed
\usepackage{amsmath} % assumes amsmath package installed
\usepackage{amssymb}  % assumes amsmath package installed
\usepackage{biblatex}
\usepackage{subcaption}
\usepackage{multirow}
\usepackage{multicol}
\usepackage{hyperref}
\usepackage{enumerate}
\usepackage{colortbl}
\usepackage{color}
\usepackage{tabularx}
\usepackage{pifont}
\usepackage[usenames,dvipsnames,table]{xcolor}
\hypersetup{
    colorlinks=true,
    linkcolor=MidnightBlue,
    filecolor=magenta,      
    urlcolor=MidnightBlue,
    citecolor=MidnightBlue,
} 
\usepackage[font=small,labelfont=bf]{caption}

\definecolor{myorange}{rgb}{1.0, 0.49, 0.0}
\NewDocumentCommand\emojirise{}{
    \includegraphics[scale=0.05]{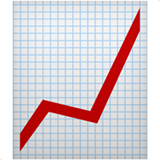}
}

\title{\LARGE \bf
RISE: 3D Perception Makes \underline{R}eal-World Robot \\\underline{I}mitation \underline{S}imple and \underline{E}ffective
}

\author{Chenxi Wang$^{1}$, Hongjie Fang$^{2}$, Hao-Shu Fang$^{2}$*, Cewu Lu$^{2}$*% <-this % stops a space
\thanks{$^{1}$ Shanghai Noematrix Intelligence Technology Ltd.}%
\thanks{$^{2}$ Shanghai Jiao Tong University.}%
\thanks{* Hao-Shu Fang and Cewu Lu are the corresponding authors.}
\thanks{Author e-mails: chenxi.wang@noematrix.cn, galaxies@sjtu.edu.cn, fhaoshu@gmail.com, lucewu@sjtu.edu.cn}
}

\begin{document}

\makeatletter
\let\@oldmaketitle\@maketitle% Store \@maketitle
\renewcommand{\@maketitle}{\@oldmaketitle% Update \@maketitle to insert...
\centering
  % \captionsetup{type=figure}
  \includegraphics[width=0.94\linewidth]
    {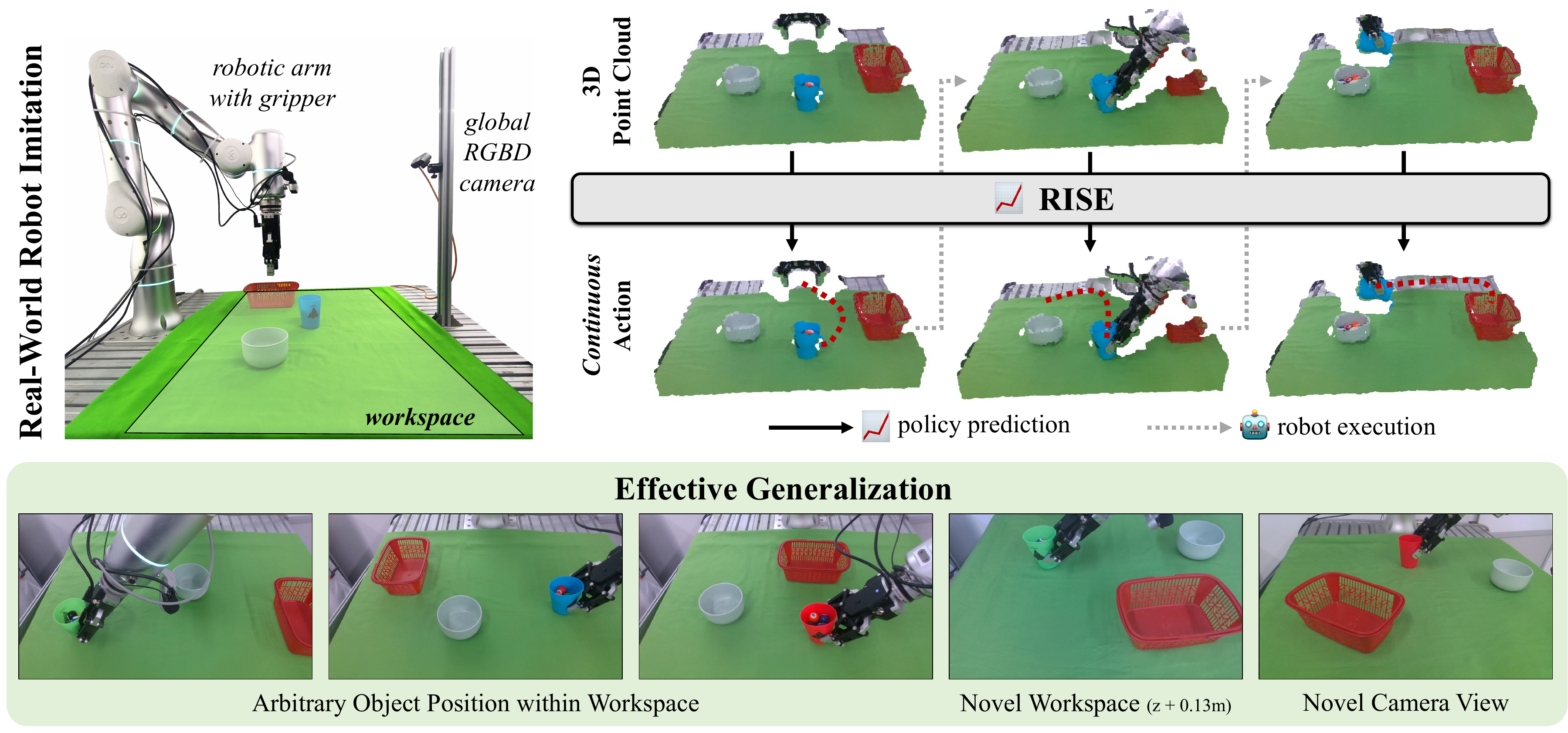}
    \captionof{figure}{\emojirise \textbf{RISE} focuses on real-world robot imitation settings with a noisy single-view partial point cloud as input, and outputs continuous robot actions. While simple, it shows effective generalization ability across object locations, novel workspaces, and novel camera views.}
    \label{fig:teaser}\vspace{-0.3cm}
    }%
\makeatother

\maketitle
\thispagestyle{empty}
\pagestyle{empty}
\addtocounter{figure}{-1}

%%%%%%%%%%%%%%%%%%%%%%%%%%%%%%%%%%%%%%%%%%%%%%%%%%%%%%%%%%%%%%%%%%%%%%%%%%%%%%%%
\begin{abstract}

Precise robot manipulations require rich spatial information in imitation learning. Image-based policies model object positions from fixed cameras, which are sensitive to camera view changes. Policies utilizing 3D point clouds usually predict keyframes rather than continuous actions, posing difficulty in dynamic and contact-rich scenarios. To utilize 3D perception efficiently, we present RISE, an end-to-end baseline for real-world imitation learning, which predicts continuous actions directly from single-view point clouds. It compresses the point cloud to tokens with a sparse 3D encoder. After adding sparse positional encoding, the tokens are featurized using a transformer. Finally, the features are decoded into robot actions by a diffusion head. Trained with 50 demonstrations for each real-world task, RISE surpasses currently representative 2D and 3D policies by a large margin, showcasing significant advantages in both accuracy and efficiency. Experiments also demonstrate that RISE is more general and robust to environmental change compared with previous baselines. Project website: \href{https://rise-policy.github.io/}{rise-policy.github.io}.

\end{abstract}

%%%%%%%%%%%%%%%%%%%%%%%%%%%%%%%%%%%%%%%%%%%%%%%%%%%%%%%%%%%%%%%%%%%%%%%%%%%%%%%%

\section{Introduction}
End-to-end policy learning framework plays an increasingly important role in robotic manipulations. With the advancements in robotics, researchers are recognizing the significance of integrating perception, planning, and execution into a continuous learning process. Recent work has made significant strides in imitation learning in an end-to-end fashion \cite{rt1,chi2023diffusion,rh20t,fang2023low,zhao2023act}. These methods allow robots to learn directly from perceptual data, bypassing the need for laborious hand-crafted feature engineering or planning steps, which opens new possibilities for addressing complex manipulation tasks and drives research and progress in the field of manipulation~\cite{rahmatizadeh2018vision}.

Spatial information is crucial for precise manipulations. For example, to pour a cup of water, the robot is required to understand the position and orientation of the cup to adjust its movements carefully and prevent water from spilling out. Image-based imitation learning tends to learn implicit spatial representations from fixed camera views \cite{rt1,chi2023diffusion,rh20t,ha2023scaling,octo,zhao2023act}. Many of these approaches utilize distinct image encoders for each view and increase the number of cameras to enhance stability and precision, consequently increasing the number of network parameters and computational overhead. In addition, such methods are susceptible to variations in camera poses, introducing a significant challenge to effective model deployment in real-world scenarios.

Spatial information can also be represented using 3D point clouds, which have demonstrated remarkable advantages in robotic tasks like general grasping \cite{anygrasp, sundermeyer2021contact}. Recently, imitation learning based on point clouds is drawing increasing interest in our community~\cite{chen2023polarnet, gervet23act3d, goyal2023rvt, guhur2022hiveformer, james2022coarse, shridar2022peract, xian2023chaineddiffuser, ze2023gnfactor, sgr23}. Most of the 3D-based methods learn to predict the next keyframe as opposed to continuous actions, which often struggle with tasks involving frequent contacts and abrupt environmental changes. Meanwhile, addressing the annotation of keyframes at scale for real-world data necessitates additional manual effort. The reason that they struggle to predict continuous actions lies in the difficulty of achieving real-time performance, which relies heavily on efficient 3D feature encoding and remains a challenge for point cloud based methods.

In this work, we propose an end-to-end imitation baseline, \textbf{RISE}, a method leveraging 3D perception to make \underline{r}eal-world robot \underline{i}mitation \underline{s}imple and \underline{e}ffective. RISE takes point clouds as input directly, and outputs continuous action trajectories for the immediate future. It employs a shallow 3D encoder built with sparse convolutions \cite{choy2019minkowski}, which effectively utilizes the advantages of conventional convolution architectures and avoids redundant computations on empty 3D spaces. Such design makes RISE able to obtain precise spatial information with a single camera, reducing the additional cost of hardware accumulation.
The encoded point features are then mapped to the action space by a transformer. Considering the unordered characteristic of 3D points, we use sparse positional encoding, a function of coordinates, to help the transformer master the relative relationships among different point tokens in 3D space. Although the point tokens are not distributed in continuous positions like language or image tokens, we can still observe stable results in scenes with variable object locations. With sparse positional encoding, the point features can be easily modeled by transformers and naturally embedded into multimodal inputs. To show the benefits of 3D perception, we only consider RISE with point cloud input in this work.
Finally, the action features are decoded into continuous trajectories by a diffusion head~\cite{chi2023diffusion}.

We test RISE in 6 real-world tasks, including \textit{pick-and-place}, \textit{6-DoF} pouring, \textit{push-to-goal} (with tool), and \textit{long-horizon} tasks. To verify the generalization ability of the policy with object locations, all the objects are randomly arranged throughout the entire workspace. Trained on 50 demonstrations for each task, RISE significantly outperforms other representative methods and keeps stable when the number of objects increases. We also find that RISE is more robust to environmental disturbance, such as changing camera views and increasing table height. Such generalization enhances the error tolerance of real-world deployment.

\section{Related Work}

\subsection{Imitation Learning for Robotics}

\begin{figure*}
    \centering
    \includegraphics[width=0.85\linewidth]{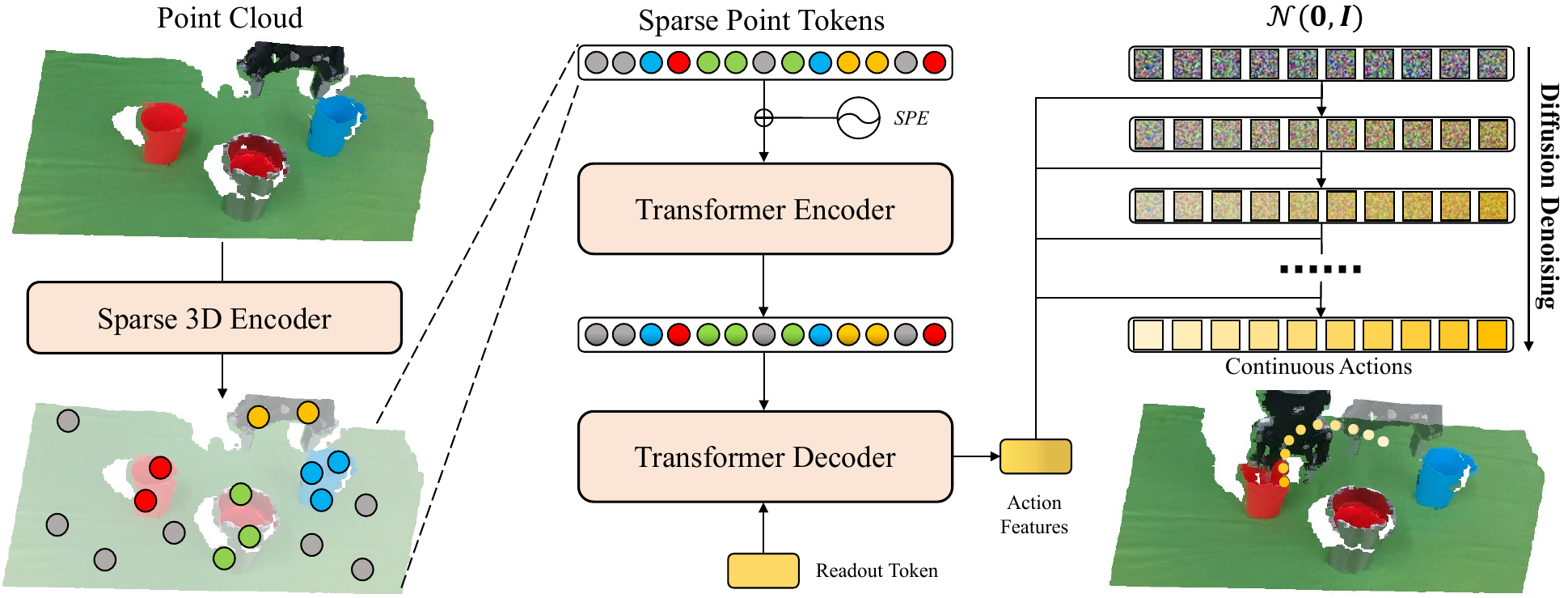}
    \caption{Overview of RISE architecture. The input of RISE is a noisy point cloud captured from the real world. A 3D encoder built with sparse convolution is employed to compress the point cloud into tokens. The tokens are fed into the transformer encoder after adding sparse positional encoding. A readout token is used to query the action features from the transformer decoder. Conditioned on the action features, the Gaussian samples are denoised into continuous actions iteratively using a diffusion head.}
    \label{fig:overview}\vspace{-0.4cm}
\end{figure*}

Imitation learning is a  machine learning paradigm where a robot learns to operate by observing and mimicking expert demonstrations. 
Behavior cloning (BC)~\cite{pomerleau1988alvinn}, as the most direct form of imitation learning, aims to identify a mapping from observations to corresponding robot actions with the supervision of the given demonstrations. 
Despite its simplicity, BC has shown promising potential in learning robotic manipulations~\cite{rt1, chi2023diffusion, jang2022bcz, mandlekar2021what, shridar2022peract, octo, zhao2023act}.

\textbf{2D Imitation Learning.} 2D image data is commonly used in imitation learning. 
One intuitive approach is to utilize pre-trained representation models for images~\cite{liv, vip, vc1, r3m, mvp} to convert them into 1D representations and map these transformed observations to the action space either through a BC policy~\cite{zhang2018deep} or non-parametric nearest neighbour~\cite{pari2021surprising}. 
Unfortunately, current pre-trained representation models are not general enough to handle diverse experimental environments, encountering trouble achieving satisfactory results in real-world settings.
Thus, many researchers learn such mapping in an end-to-end manner~\cite{rt1, chi2023diffusion, jang2022bcz, mandlekar2021what, oxe, gato, octo, zhao2023act, rt2} and have demonstrated impressive performance across many tasks. 
Specifically, ACT~\cite{zhao2023act} adopts a CVAE scheme~\cite{sohn2015learning} with transformer backbones~\cite{vaswani2017attention} and ResNet image encoders~\cite{he2016resnet} to model the variability of human data, while Diffusion Policy~\cite{chi2023diffusion} directly utilizes diffusion process~\cite{ho2020denoising} to express multimodal action distributions generatively. %%% TODO
Nonetheless, these policies are sensitive to camera positions and often fail to capture 3D spatial information about the objects in the environments. 

\textbf{3D Imitation Learning.} 
The formulation of incorporating 3D information in the imitation learning framework is under active exploration. 
The most straightforward method is to apply projections to transform the 3D point cloud to several 2D image views and transfer the task to multi-view image-based policy learning~\cite{goyal2023rvt,guhur2022hiveformer}, which requires the virtual viewpoints to be carefully designed to ensure performance. 
Moreover, due to sparse and noisily sensed point clouds, \cite{goyal2023rvt} fails to grasp slim objects like marker pens in real-world experiments. 
\cite{james2022coarse,shridar2022peract,ze2023gnfactor} process point clouds to dense voxel grids and apply 3D convolutions. 
Since high-resolution 3D feature maps require expensive computes, these methods have to trade off performance against cost. \cite{zhu2023learning} proposed an object-centric representation for learning which requires an additional segmentation process.
\cite{gervet23act3d,xian2023chaineddiffuser} featurize point clouds by projecting multi-view image features to 3D world to avoid dense convolutions. 
However, such feature fusion techniques struggle to capture the consistent 3D representation from different views accurately. 
Recently, a concurrent work DP3~\cite{ze2024dp3} also leverages 3D perception in robotic manipulation policies, but our real-world evaluations in \S\ref{sec:exp-ablations} demonstrate that it cannot handle demonstrations with various representations limited by its network capacity. DexCap~\cite{wang2024dexcap} also proposed a PointNet~\cite{qi2017pointnet} with diffusion head architecture for dexterous manipulation.

As mentioned before, most of the current 3D robotic imitation learning methods predict keyframes instead of continuous action, which makes it hard to annotate and limits their capacity. Besides, many of these methods only show results in simulation environments like RLBench~\cite{james2020rlbench} and CALVIN~\cite{mees2022calvin}. In this work, we aim to evaluate our method in a more challenging setting: continuous action control with a noisy single-view partial point cloud in the real world.

\subsection{3D Perception}

3D perception has received considerable attention from researchers in the computer vision and robotics communities. It can be roughly divided into the following three categories: 

\textbf{Projection-based.} This approach initially projects the 3D point cloud onto multiple images on different planes and then employs traditional multi-view image perception techniques. It is widely applied in shape recognition~\cite{hamdi2021mvtn}, object detection~\cite{chen2017multi, li2016vehicle} and robotic manipulations~\cite{goyal2023rvt, guhur2022hiveformer} due to its simplicity. However, the projections can lead to the geometric information loss of the 3D data, and the sensitivity to the choice of projection planes may result in inferior performance~\cite{zhao2021point}.

\textbf{Point-based.} Early researchers directly utilized 3D convolutional neural networks (CNNs) to process 3D point cloud data based on dense volumetric representations~\cite{dai2017scannet, wu20153d, zhou2018voxelnet}. Still, the sparsity of 3D data makes the vanilla approaches inefficient and memory-intensive. To solve this problem, researchers have explored using octrees for memory footprint reduction~\cite{riegler2017octnet, wang2017cnn}, utilizing sparse convolutions to minimize unnecessary computations in inactive regions to improve efficiency and effectiveness~\cite{choy2019minkowski, graham20183d}, and aggregating features across point sets directly using different network architectures~\cite{pan20213d, qi2017pointnet, qian2022pointnext, zhao2021point}.

\textbf{NeRF-based.} Neural radiance fields (NeRFs)~\cite{mildenhall2021nerf} have demonstrated impressive performance on high-fidelity 3D scene synthesis and scene representation extractions. In recent years, some studies~\cite{driess2022reinforcement, shen2023distilled, ye2023featurenerf, ze2023gnfactor} have employed features extracted from pre-trained 2D foundational models as additional supervisory signals in NeRF training for scene feature extraction and distillation. Nevertheless, NeRF training requires image data from multiple views, which poses obstacles for scaling up in real-world environments. Additionally, it does not align with our single-view setting.

\section{Method}

Given a point cloud $\mathcal{O}^t=\{P_i^t=(x_i^t,y_i^t,z_i^t,r_i^t,g_i^t,b_i^t)\}$ as the observation at time $t$, RISE aims to predict the next $n$-step robot actions $\mathcal{A}^t=\{A_{t+1}, A_{t+2}, \cdots, A_{t+n}\}$, where $A_i$ contains the translation, rotation and width of the gripper.
Due to the large domain gap between point clouds and robot actions, it is challenging to learn the approximation $f: \mathcal{O}^t \rightarrow \mathcal{A}^t$ directly. To model the process, RISE is decomposed into three functions: a sparse 3D encoder $h_\mathrm{E}: \mathcal{O}^t \rightarrow \mathcal{F}_\mathrm{P}^t$, a transformer $h_\mathrm{T}: \mathcal{F}_\mathrm{P}^t \rightarrow \mathcal{F}_\mathrm{A}^t $ and an action decoder $h_\mathrm{D}: \mathcal{F}_\mathrm{A}^t \rightarrow \mathcal{A}^t$, where $\mathcal{F}_\mathrm{P}^t$ and $\mathcal{F}_\mathrm{A}^t$ denote the features of point clouds and actions respectively.

\subsection{Modeling Point Clouds using Sparse 3D Encoder}
The most significant difference between point cloud data and images is that point clouds are sparse and unorganized, which makes CNNs unsuitable to be applied to the points. For inputs at different scales, the computation efficiency and flexibility of a model should be taken into consideration.
We employ a 3D encoder built on sparse convolution \cite{choy2019minkowski}. It keeps most of the standard convolution, while only computes outputs on predefined coordinates. Such an operator saves computation and inherits the core advantage of conventional convolution.

The sparse 3D encoder $h_\mathrm{E}$ adopts a shallow ResNet architecture \cite{he2016resnet}. It is composed of one initial convolution layer, four residual blocks, and one final convolution layer, with five $2\times$ sparse pooling layers between every two components. The number of layers can be freely increased, while the evaluation results demonstrate that a shallow encoder is sufficient for our experiments.

By $h_\mathrm{E}$, the voxelized point cloud $\mathcal{O}^t$ is encoded to sparse point features $\mathcal{F}_\mathrm{P}^t$ in an efficient way, avoiding redundant computing on huge empty space. $\mathcal{F}_\mathrm{P}^t$ is then fed into the transformer $h_\mathrm{T}$ as sparse tokens. For $\mathcal{O}^t$ cropped in $1\times 1\times 1\mathrm{m}^3$ space, $\mathcal{F}_\mathrm{P}^t$ contains only 60 $\sim$ 80 tokens. Although the token number is less than the one in ACT~\cite{zhao2023act} (300 per image), experiments in \S\ref{sec:exp-ablations} show that point cloud based ACT still outperforms the original implementation.

\subsection{Transformer with Sparse Point Tokens}

\begin{figure*}
\vspace{0.1cm}
    \centering
    \includegraphics[width=0.95\linewidth]{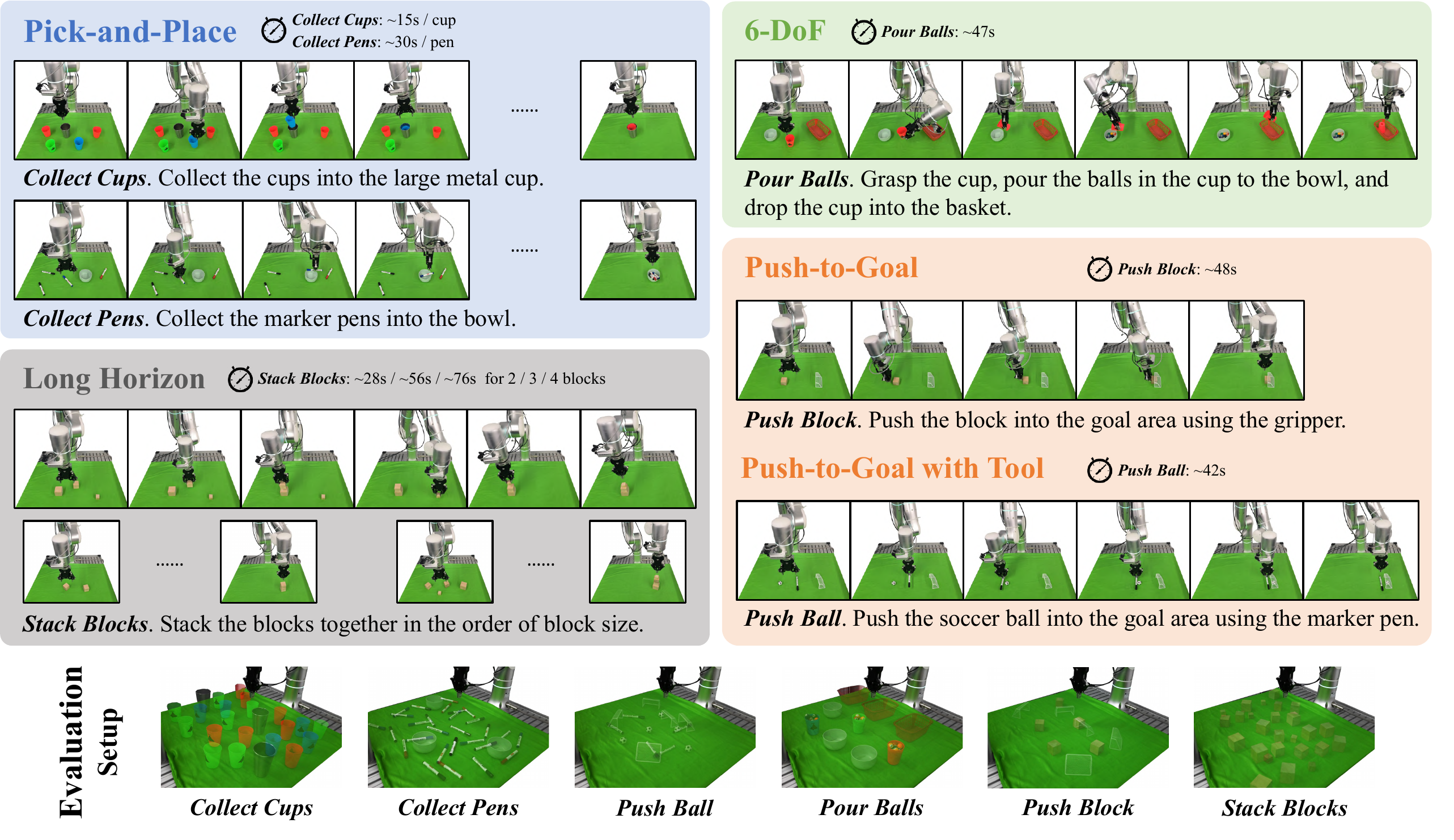}
    \caption{Definition of the tasks in the experiments. During evaluation, each task is randomly initialized within the robot workspace. For each task, only 3 to 5 setups from the evaluations are depicted in the figure for clarity.}
    \label{fig:tasks}\vspace{-0.4cm}
\end{figure*}

We adopt transformer~\cite{vaswani2017attention} to implement the mapping from point features $\mathcal{F}_\mathrm{P}^t$ to action features $\mathcal{F}_\mathrm{A}^t$. While the positional encoding for image tokens is dense and natural, sparse point tokens could not be processed in the same manner. We instead introduce sparse positional encoding for point tokens.

Let $(x,y,z)$ be the coordinate of the point token $P$ with $d$-dimension, the position of $P$ is defined as
\begin{equation}
    \begin{split}
        &pos_k = \frac{k}{v}+c, \ \ k\in\{x,y,z\},\\
        &pos = [pos_x, pos_y, pos_z],
    \end{split}
\end{equation}
where $c$ and $v$ are fixed offsets, and $[\cdot]$ stands for vector concatenation. The encoding dimension along each axis is $d_x=d_y=\lfloor d/3\rfloor$, $d_z=d-d_x-d_y$. The position encoding of $P$ is computed by $SPE = [SPE^x, SPE^y, SPE^z]$ where
\begin{equation}
    \left\{
    \begin{aligned}
        &SPE^k_{(pos,2i)} = \sin \frac{pos_k}{10000^{2i/d_k}}\\
        &SPE^k_{(pos,2i+1)} = \cos \frac{pos_k}{10000^{2i/d_k}}
    \end{aligned}
    \right.,\quad k\in\{x,y,z\}
\end{equation}

With the help of sparse positional encoding, we effectively capture intricate 3D spatial relationships among unordered points, which enables seamless embedding of the 3D features into conventional transformers.

The transformer $h_\mathrm{T}$ utilizes an encoder-decoder architecture, taking point features $\mathcal{F}_\mathrm{P}^t$ as input tokens without other proprioceptive signals. In the transformer decoding step, we use one readout token to query action features $\mathcal{F}_\mathrm{A}^t$.

\subsection{Diffusion as Action Decoder}
The action decoder $h_D$ is implemented as a denoising process by diffusion~\cite{chi2023diffusion, ho2020denoising, janner2022planning}. Conditioning on $\mathcal{F}_\mathrm{A}^t$, $h_\mathrm{D}$ denoises the Gaussian noises $\mathcal{N}(0, \sigma^2 I)$ to actions $\mathcal{A}^t$ iteratively. The denoising process of step $k$ is
\begin{equation}
    \mathcal{A}_{k-1}^t = \alpha (\mathcal{A}_k^t - \gamma\epsilon_\theta(\mathcal{O}^t,\mathcal{A}_k^t,k) + \mathcal{N}(0, \sigma^2 I)),
\end{equation}
where $\epsilon_\theta$ is a network predicting noises with parameters $\theta$, $\alpha$, $\gamma$ and $\sigma$ are hyperparameters related to $k$ in noise schedule. The objective function is the simplified objective in~\cite{ho2020denoising}. We use the DDIM scheduler \cite{song21denoising} to accelerate the inference speed in real-world experiments.

The regression head is also frequently employed due to its simplicity \cite{gervet23act3d,guhur2022hiveformer,jang2022bcz,zhao2023act}, whereas the diffusion head excels in handling scenes with multiple targets. Moreover, diffusion produces diverse trajectories to the same target, as opposed to averaging learned trajectories \cite{chi2023diffusion}.

For all tasks in our experiments, RISE adopts a unified action representation in the camera coordinate system, which is composed of translations, rotations, and gripper widths. We opt for absolute position for translation and 6D representation \cite{rot6d} for rotation with continuity considerations.

\section{Experiments}

\subsection{Setup}

In the experiments, we use a Flexiv Rizon robotic arm\footnote{\url{https://www.flexiv.com/product/rizon}} equipped with a Dahuan AG-95 gripper\footnote{\url{https://en.dh-robotics.com/product/ag}} for interacting with objects. Two Intel RealSense D435 RGBD cameras\footnote{\url{https://www.intelrealsense.com/depth-camera-d435}} are installed for scene perception. One global camera is positioned in front of the robot, while the other in-hand camera is mounted on the end-effector of the arm. 
For 3D perception, only the global camera is used to generate a noisy single-view partial point cloud; while for image-based policies, both cameras are used for a better understanding of spatial geometries.
All devices are linked to a workstation with an Intel Core i9-10900K CPU and an NVIDIA RTX 3090 GPU for both data collection and evaluation. 

We carefully designed 6 tasks from 4 types for the experiments as illustrated in Fig.~\ref{fig:tasks}: the \textit{pick-and-place} tasks (\textbf{\textit{Collect Cups}} and \textbf{\textit{Collect Pens}}), the \textit{6-DoF} tasks (\textbf{\textit{Pour Balls}}), the \textit{push-to-goal} tasks (\textbf{\textit{Push Block}} and \textbf{\textit{Push Ball}}) and the \textit{long-horizon} tasks (\textit{\textbf{Stack Blocks}}).
The data collection setup and process are the same as~\cite{rh20t}, \textit{i.e.}, end-effector tele-operations using a haptic device. Unless specifically stated, we gathered 50 expert demonstrations for each task as the training data for the policy, and each policy was tested for 20 consecutive trials to evaluate its performance. During evaluations, objects in the task are randomly initialized within the robot workspace of approximately $50\text{cm}\times 70\text{cm}$.

\subsection{Pick-and-Place Tasks}

As the cornerstone of robotic manipulations, \textit{pick-and-place} tasks emphasize precise manipulations of objects and efficient generalizations of the robot policy within the workspace. As shown in Fig.~\ref{fig:tasks}, the \textbf{\textit{Collect Cups}} task is designed to evaluate the policy performance in predicting the translation part of the action, whereas the \textbf{\textit{Collect Pens}} task is utilized to further assess the ability of the policy in predicting the planner rotation part of the action.

We employ two representative image-based policies as our baselines: ACT~\cite{zhao2023act} and Diffusion Policy~\cite{chi2023diffusion}. We also evaluate a keyframe-based 3D policy Act3D~\cite{gervet23act3d}, the current state-of-the-art policy on RLBench~\cite{james2020rlbench} in the experiments. For Act3D, we utilize a simple action planner to execute the predicted keyposes, preventing collisions with other objects in the workspace.

For each of the five scenarios where there are 1, 2, 3, 4, and 5 objects in the workspace, we collected 10 demonstrations, resulting in a total of 50 demonstrations used for the policy training. During evaluations, we conducted 10 trials for each of the five scenarios, tallying the number of cups placed into the large metal cup or the number of pens placed into the bowl, and calculating the completion rate. For each object, we imposed a runtime limit of 20 keyframes for keyframe-based policies and 300 steps for continuous-control policies.

\begin{figure}[h]
\centering
\vspace{-0.4cm}
\begin{subfigure}[b]{0.48\linewidth}
    \includegraphics[width=\linewidth]{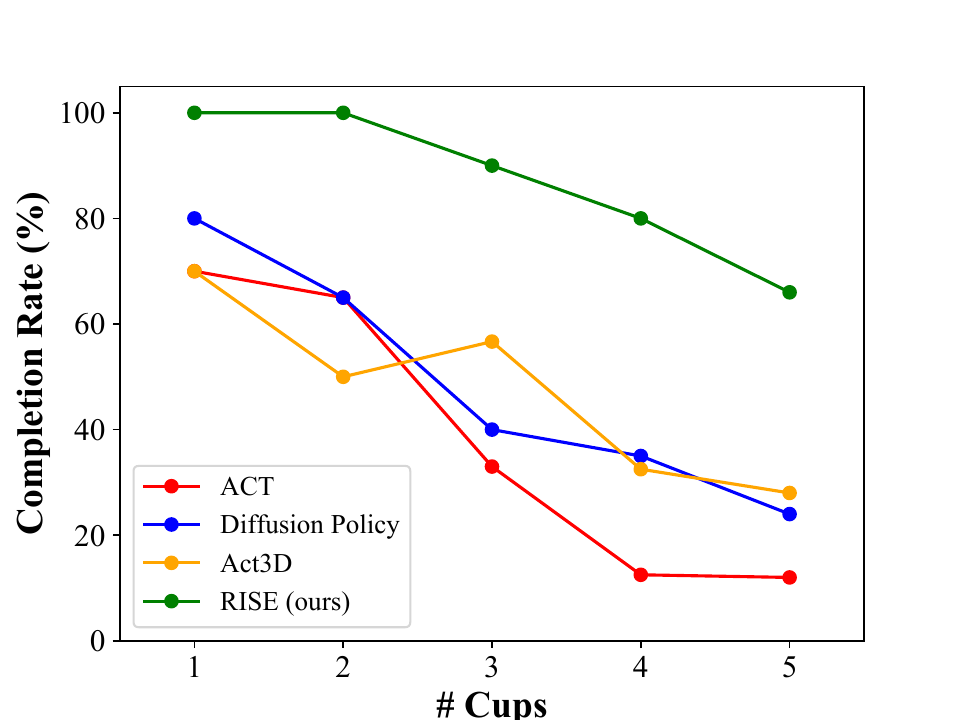}
    \caption{\textbf{\textit{Collect Cups}}}\label{fig:exp-collect-cups}
\end{subfigure}
\begin{subfigure}[b]{0.48\linewidth}
    \includegraphics[width=\linewidth]{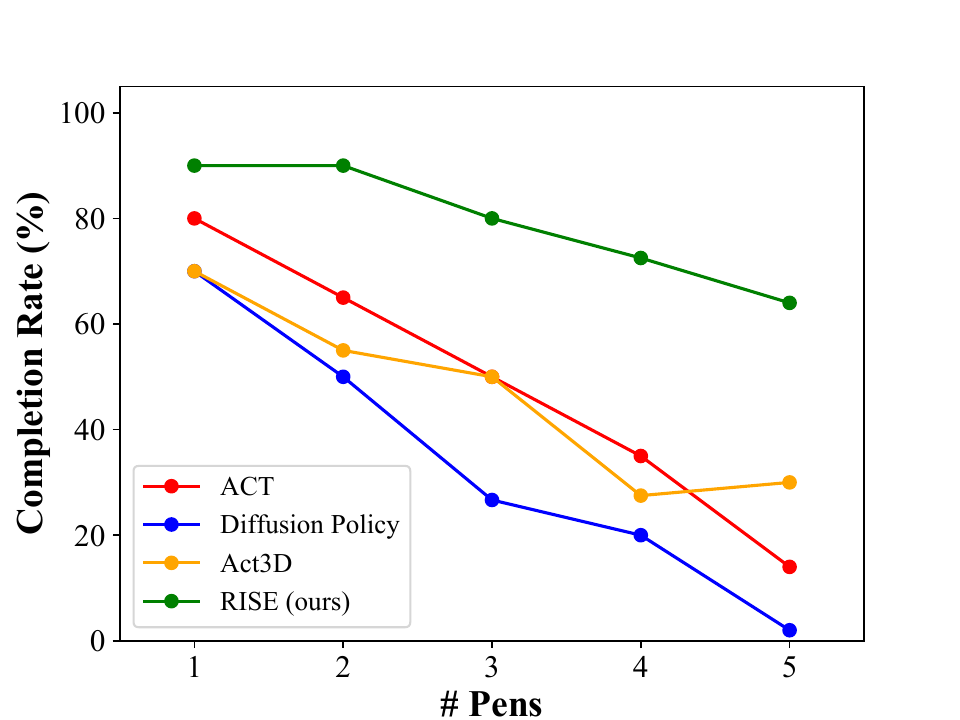}
    \caption{\textbf{\textit{Collect Pens}}}\label{fig:exp-collect-pens}
\end{subfigure}
    \caption{Experimental results of the \textit{pick-and-place} tasks.}\label{fig:exp-pick-and-place}
\vspace{-0.2cm}
\end{figure}

The evaluation results are depicted in Fig.~\ref{fig:exp-pick-and-place}.
In the \textbf{\textit{Collect Cups}} task, RISE achieves a completion rate of over 90\% when the number of cups is less than 3. Even in complex environments with 4 or 5 cups, RISE maintains a completion rate of over 65\%, showcasing capabilities surpassing all baselines. Similar trends are observed in the \textbf{\textit{Collect Pens}} task: RISE consistently outperforms all baselines, demonstrating its ability to not only predict the translation part but also accurately forecast planner rotation. We also discover that Act3D performs comparably to image-based baselines. Moreover, given that Act3D requires specially designed motion planners for more complicated actions and cannot provide immediate responses to sudden changes in the environment, we therefore only employ ACT and Diffusion Policy as baselines in our subsequent experiments.

\subsection{6-DoF Tasks}

The \textit{6-DoF} \textbf{\textit{Pour Balls}} task is designed to test the ability of robot policies to forecast actions with complex spatial rotations, instead of the simple planner rotation in the \textit{pick-and-place} tasks. As shown in Fig.~\ref{fig:tasks}, the robotic arm needs to undergo complex spatial rotations to complete the task, at times approaching its kinematic limits. During evaluations, 10 balls are initialized in the cup. Besides the action success rates of the policy, the number of balls poured into the cup is also recorded to calculate the completion rates. The runtime limit for this task is set as 1200 steps.

The experimental results are shown in Tab.~\ref{tab:pour-balls}. From the action success rates, it is evident that RISE can learn actions with complex spatial rotations more effectively compared to image-based policies. Additionally, its execution of the pouring action is more precise regarding pouring positions, resulting in higher task completion rates. This also highlights the effectiveness of 3D perception, which can capture more accurate spatial relationships between objects.

\begin{table}[h]
    \centering
    \setlength\tabcolsep{4pt}
    \renewcommand{\arraystretch}{1.2}
    \begin{tabular}{ccccccc}\hline
        \multirow{2}{*}{\textbf{Method}} & 
        \multicolumn{3}{c}{\textbf{Success Rate (\%)}} & &
        \multicolumn{2}{c}{\textbf{Completion Rate (\%)}} 
        \\ 
        \cline{2-4}\cline{6-7}   
        & \textbf{Grasp} &\textbf{Pour}  & \textbf{Place} & & \textbf{Overall} & \textbf{If Poured}\\\hline
        ACT~\cite{zhao2023act} & 30 & 30 & 0 && 13.0 & 43.3\\
        Diffusion Policy~\cite{chi2023diffusion} & 55 & 55 & 35 && 30.5 & 55.5 \\
        RISE (ours) & \textbf{80} & \textbf{80} & \textbf{70} && \textbf{49.0} & \textbf{61.3}
        \\\hline
    \end{tabular}
    \caption{Experimental results of the \textit{6-DoF} task \textbf{\textit{Pour Balls}}. ``If Poured'' denotes the completion rate in the context of having completed the pouring action.}
    \label{tab:pour-balls}\vspace{-0.4cm}
\end{table}

\subsection{Push-to-Goal Tasks}

\begin{figure*}
\centering
    \begin{minipage}{0.25\linewidth}
        \includegraphics[width=\linewidth]{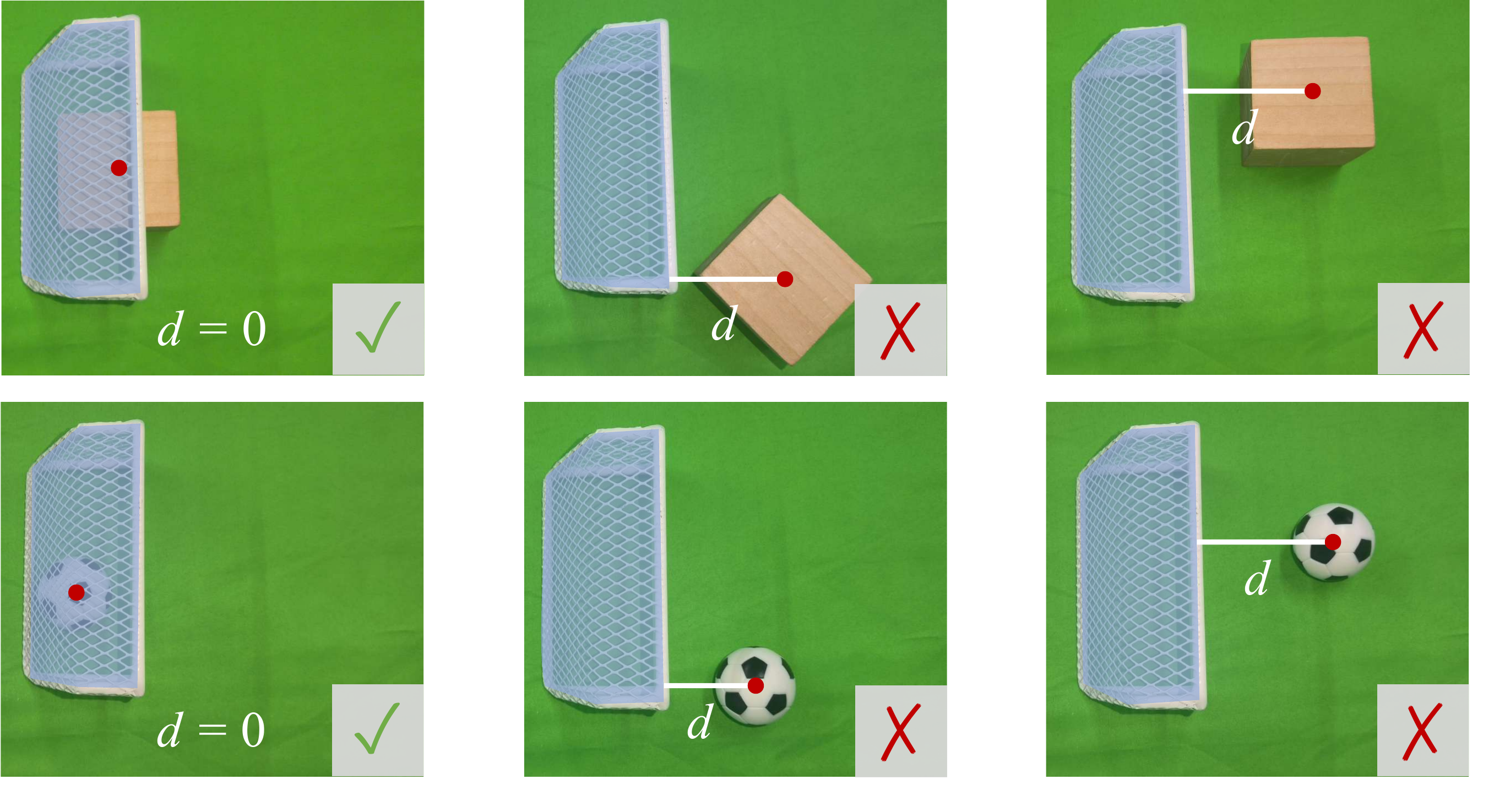} 
    \end{minipage}
    \begin{minipage}{0.65\linewidth}
        \centering
        \footnotesize
        \setlength\tabcolsep{4pt}
        \renewcommand{\arraystretch}{1.2}
        \begin{tabular}{cccc}\hline
            \textbf{Task} & \textbf{Method} & \textbf{Success Rate (\%)} & \textbf{Average Distance $d$ (cm)} $\downarrow$ \\ \hline
            \multirow{2}{*}{\textit{\textbf{Push Block}}} & Diffusion Policy~\cite{chi2023diffusion} & 50 & 5.67 \\ 
            & RISE (ours) & \textbf{55} & \textbf{3.51} \\ \hline
            \multirow{2}{*}{\textit{\textbf{Push Ball}}} & Diffusion Policy~\cite{chi2023diffusion} & 30 & 6.05 \\
            & RISE (ours) & \textbf{60} & \textbf{4.89}
            \\\hline 
        \end{tabular} 
    \end{minipage}
    \captionof{table}{Evaluation metrics illustrations (left) and experimental results (right) of the \textit{push-to-goal} tasks \textbf{\textit{Push Block}} and \textbf{\textit{Push Ball}}.}
    \label{tab:push-to-goal}\vspace{-0.5cm}
\end{figure*}

Robot policies should generate immediate feedback to environmental dynamics, enabling adaptation to object movements within the environment to accomplish tasks. To this end, we designed two \textit{push-to-goal} tasks \textbf{\textit{Push Block}} and \textbf{\textit{Push Ball}}, as shown in Fig.~\ref{fig:tasks}. Furthermore, the \textbf{\textit{Push Ball}} task requires the robot to use a tool (a marker pen) to complete the task, thereby testing the ability of the policy to utilize tools. In the evaluation process, we compute the distance $d$ from the object center to the goal area. If the object center is within the goal area ($d=0$), it is considered a success, as shown in Tab.~\ref{tab:push-to-goal} (left). We then calculate the task success rate and the average distance as metrics. The runtime limit for this task is set as 1200 steps.

The evaluation results are shown in Tab.~\ref{tab:push-to-goal} (right). In the \textbf{\textit{Push Block}} task, RISE slightly surpasses Diffusion Policy in terms of success rate, while pushing the block closer to the goal area. 
However, in the \textbf{\textit{Push Ball}} task, RISE outperforms Diffusion Policy by a significant margin, demonstrating its effective 3D perception of object position changes and ability to rapidly adjust policy action outputs accordingly. 
We also observe that ACT struggles in both tasks and frequently causes the robot to make hard contact with the objects and triggers the mechanical emergency stop, which might be attributed to its imprecision in scene perception.

\subsection{Long-Horizon Tasks}

Long-horizon tasks are crucial in robotic manipulations as they highlight the impact of error accumulation in actions over extended horizons, providing insights into the robustness and adaptability of the policy.
Therefore, we designed the \textit{long-horizon} task \textbf{\textit{Stack Blocks}} to assess the policy's ability in this regard, given that blocks are more likely to topple as the stack grows. We emphasize that this task is more challenging than previous \textit{pick-and-place} tasks because (a) some blocks are only slightly smaller than gripper width, requiring precise controls in grasping; (b) the policy needs to recognize the sizes of the blocks and select a suitable stacking order to ensure the stability of the resulting stack; and (c) as the stack grows, the policy needs to dynamically adjust the placement height of the blocks to ensure that they do not collide with existing blocks and can be smoothly placed on top of them.

For three cases where there are 2, 3, and 4 blocks in the workspace, we collected 10, 20, and 20 demonstrations respectively, totaling 50 demonstrations for the policy training. During evaluations, 10 trials are conducted for each case, and the average number of successfully stacked blocks is reported to measure the policy performance. We set the runtime limit for the task as 600, 1200, and 1800 steps for each case respectively.

\begin{table}[h]
\centering
\footnotesize
\setlength\tabcolsep{4pt}
\renewcommand{\arraystretch}{1.2}
\begin{tabular}{cccc} \hline
     \multirow{2}{*}{\textbf{Method}} & \multicolumn{3}{c}{\textbf{\# Blocks}} 
     \\ \cline{2-4}  & \textbf{2} & \textbf{3} & \textbf{4}  \\
     \hline
     ACT~\cite{zhao2023act} & 0.6 / 1 & 0.5 / 2 & 0.3 / 3\\
     Diffusion Policy~\cite{chi2023diffusion} & 0.7 / 1 & 0.5 / 2 & 0.5 / 3 \\
     RISE (ours) & \textbf{0.8} / 1 & \textbf{1.5} / 2 & \textbf{0.9} / 3\\
     \hline
\end{tabular}
\caption{Experimental results (average stacked blocks) of the \textit{long-horizon} task \textbf{\textit{Stack Blocks}}.}
\label{tab:stack-block}\vspace{-0.2cm}
\end{table}

The experimental results are shown in Tab.~\ref{tab:stack-block}. In the simple scenario with only two blocks, all policies yielded similar results; however, as the number of blocks increased, RISE progressively surpassed the baselines significantly, showcasing its ability to adapt well to \textit{long-horizon} tasks and control accumulated errors. Moreover, we observe that the baselines exhibit a higher frequency of the aforementioned issues (b) and (c) compared to RISE, implying that powered by 3D perception, RISE has a deeper understanding of the scene, and its predictions of actions are also more precise.

\subsection{Effectiveness of 3D Perception}\label{sec:exp-ablations}

In this section, we explore how 3D perception enhances the performance of robot manipulation policies on the \textbf{\textit{Collect Cups}} task with 5 cups. We replace the image encoder of the image-based policies ACT and Diffusion Policy with the sparse 3D encoder used in RISE. The experiment results are shown in Tab.~\ref{tab:3d}. We observe a significant improvement in the performance of ACT and Diffusion Policy after applying 3D perception even with fewer camera views, surpassing the 3D policy Act3D, which reflects the effectiveness of our 3D perception module in manipulation policies.

\begin{table}[t]
\vspace{0.2cm}
\definecolor{darkpastelred}{rgb}{0.76, 0.23, 0.13}
\centering
        \footnotesize
        \setlength\tabcolsep{4pt}
        \renewcommand{\arraystretch}{1.2}
        \begin{tabular}{cccl}\hline
             \textbf{Method} & \textbf{3D} & \textbf{\# Cameras} & \textbf{Completion Rate (\%)} \\ \hline
             \multirow{2}{*}{ACT~\cite{zhao2023act}} & & 2 & \quad \quad \quad 12 \\
             & \checkmark & 1 & \quad \quad \quad 32{ \color{darkpastelred}$^{\uparrow 20}$} \\ \hline
             \multirow{2}{*}{Diffusion Policy~\cite{chi2023diffusion}} & & 2 &  \quad \quad \quad 24\\
             & \checkmark & 1 & \quad \quad \quad 36{ \color{darkpastelred}$^{\uparrow 12}$}\\ \hline
             DP3*~\cite{ze2024dp3} & \checkmark & 1 &  \quad \quad \quad \space - \\
             Act3D~\cite{gervet23act3d} & \checkmark & 1 & \quad \quad \quad 28\\
             
             RISE (ours) & \checkmark & 1 & \quad \quad \quad \textbf{66}
            \\\hline 
        \end{tabular} 
        \caption{Effectiveness test of 3D perception on the \textbf{\textit{Collect Cups}} task with 5 cups (10 trials). 2D version of policies take images from both global and in-hand cameras as input. * DP3 fails to learn in our setting, see text for a more detailed analysis.}\label{tab:3d}\vspace{-0.2cm}
\end{table}

\begin{figure}[t]
    \begin{minipage}{0.45\linewidth}
        \centering
        \includegraphics[width=\linewidth]{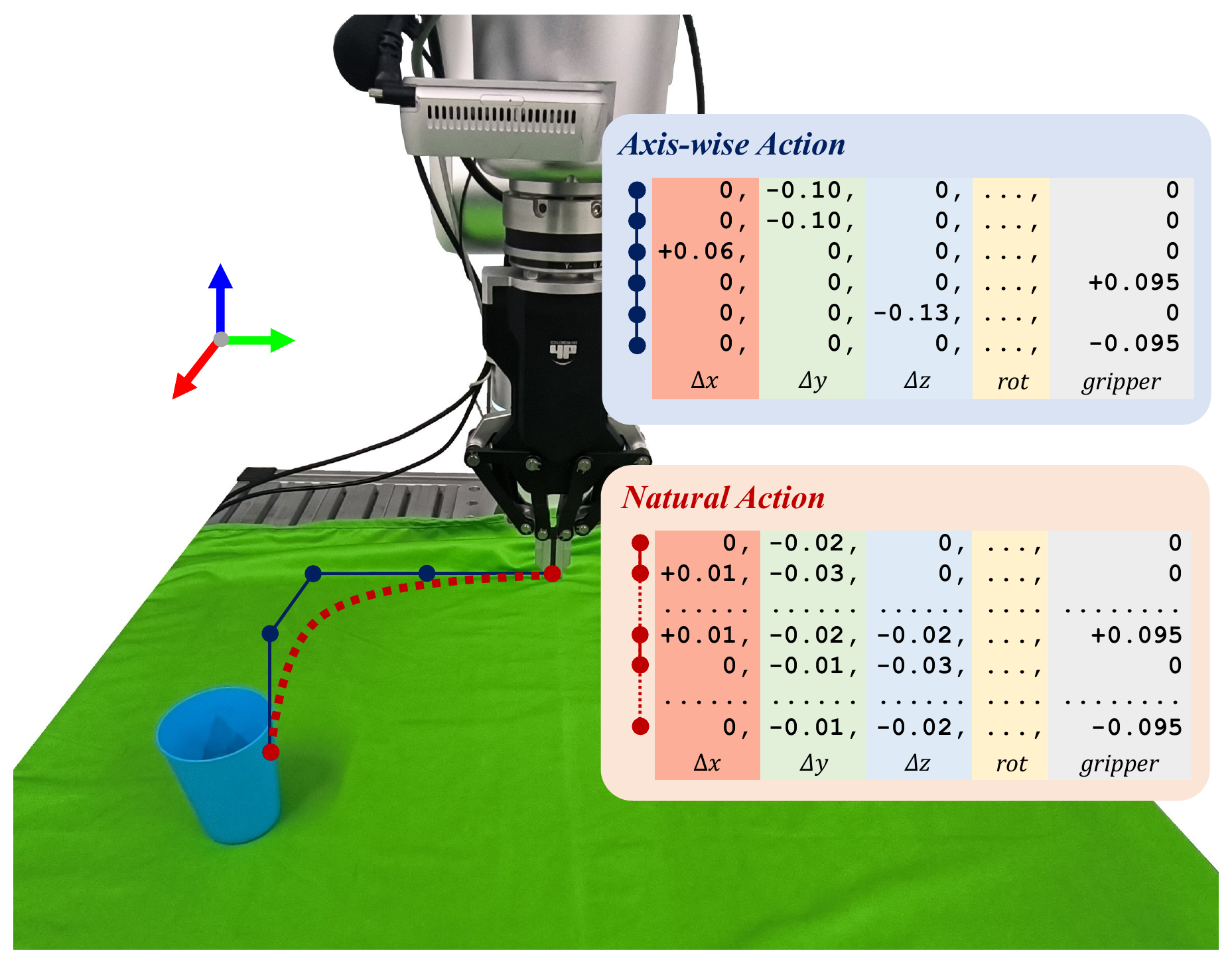} 
    \end{minipage}\hspace{0.1cm}
    \begin{minipage}{0.45\linewidth}
    \centering\footnotesize
            \setlength\tabcolsep{0.8pt}
            \renewcommand{\arraystretch}{1.2}
            \begin{tabular}{ccc}
            \hline
            \multirow{2}{*}{\textbf{Method}} & \multicolumn{2}{c}{\textbf{Completion Rate (\%)}} \\ \cline{2-3} 
            & Axis-wise & Natural \\
            \hline
            DP3, hor. 4 & 0 & 0\\
            DP3, hor. 8 & 0 & 0\\
            DP3, hor. 16 & 20 & 0\\
            DP3, hor. 24 & 40 & 0\\ \hline
            RISE & \textbf{80} & \textbf{100} \\ \hline
            \end{tabular}
    \end{minipage}
    
    \hspace{0.04\linewidth}

    \captionof{table}{Analysis of the failures of DP3 in our experiments. (left) Illustrations of the axis-wise and natural action. (right) Results of the \textbf{\textit{Collect Cups}} task with 1 cup when using demonstrations with different action representations for training (10 trials).}\label{tab:exp-dp3}\vspace{-0.4cm}
\end{figure}

\begin{figure*}
\centering
    \begin{minipage}{0.45\linewidth}
\includegraphics[width=\linewidth]{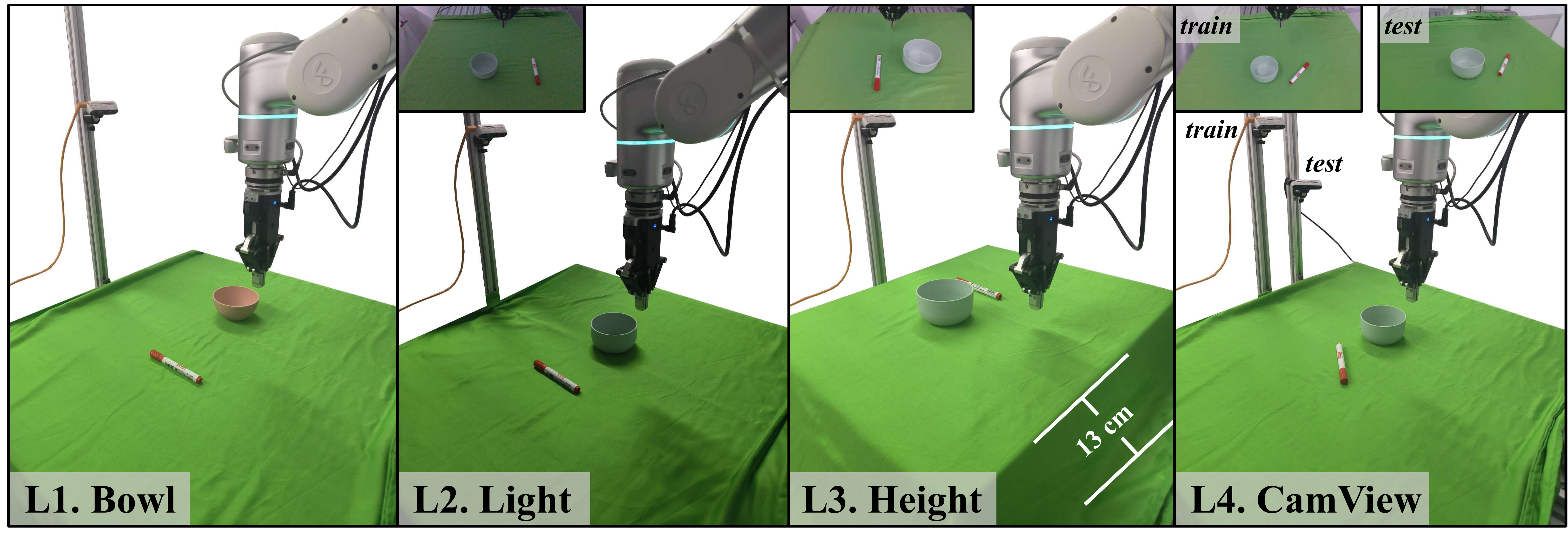} 
    \end{minipage}
    \hspace{0.04\linewidth}
    \begin{minipage}{0.45\linewidth}
    \definecolor{darkpastelgreen}{rgb}{0.01, 0.75, 0.24}
    \definecolor{darkgray}{rgb}{0.66, 0.66, 0.66}
    \centering
            \footnotesize
            \setlength\tabcolsep{3pt}
            \renewcommand{\arraystretch}{1.2}
            \begin{tabular}{cccccc}\hline
                 \multirow{3}{*}{\textbf{Method}} & \multicolumn{5}{c}{\textbf{Completion Rate (\%)}} \\ \cline{2-6}
                 & \multirow{2}{*}{\textbf{Original}} & \multicolumn{4}{c}{\textbf{Disturbance}} \\ \cline{3-6}
                 & & \textbf{Bowl} & \textbf{Light} & \textbf{Height} & \textbf{CamView} \\ \hline
                 ACT~\cite{zhao2023act} & 80 & 70{ \color{darkpastelgreen}$^{\downarrow \underline{10}}$} & 40{ \color{darkpastelgreen}$^{\downarrow 40}$} & 0{ \color{darkpastelgreen}$^{\downarrow 80}$} & 0{ \color{darkpastelgreen}$^{\downarrow 80}$}\\
                 Diffusion Policy~\cite{chi2023diffusion} & 70 &  50{ \color{darkpastelgreen}$^{\downarrow 20}$} & 30{ \color{darkpastelgreen}$^{\downarrow 40}$} & 0{ \color{darkpastelgreen}$^{\downarrow 70}$} & 0{ \color{darkpastelgreen}$^{\downarrow 70}$}\\
                 Act3D~\cite{gervet23act3d} & 70 & 40{ \color{darkpastelgreen}$^{\downarrow 30}$} & 60{ \color{darkpastelgreen}$^{\downarrow \underline{10}}$} & 50{ \color{darkpastelgreen}$^{\downarrow \underline{20}}$} & 10{ \color{darkpastelgreen}$^{\downarrow 60}$}\\
                 RISE (ours) & \textbf{90} & \textbf{80}{ \color{darkpastelgreen}$^{\downarrow \underline{10}}$} & \textbf{80}{ \color{darkpastelgreen}$^{\downarrow \underline{10}}$} & \textbf{80}{ \color{darkpastelgreen}$^{\downarrow \underline{10}}$} & \textbf{50}{ \color{darkpastelgreen}$^{\downarrow \underline{40}}$}
                \\\hline 
            \end{tabular} 
        \end{minipage}
        \captionof{table}{Generalization test setup and experimental results of the \textbf{\textit{Collect Pens}} task with 1 pen (10 trials).}\label{tab:exp-general}\vspace{-0.5cm}
\end{figure*}

We also evaluate the recently proposed DP3~\cite{ze2024dp3} in this experimental setting. However, DP3 appears to struggle in learning meaningful actions from our demonstration data. After communicating with the authors, one potential reason is that they are using RealSense L515 in their original experiments, and we adopted RealSense D435 in our experiments. The point cloud from D435 is noisier, making it more challenging for networks to learn. By using sparse convolution, RISE is more robust to the noise in the point cloud. Besides, after delving into their real robot experiments, we found that instead of natural actions, axis-wise actions are used in their demonstration data, as illustrated in Tab.~\ref{tab:exp-dp3} (left). Hence, we collect 50 demonstrations on the \textbf{\textit{Collect Cups}} task with 1 cup using axis-wise and natural action representations respectively. These demonstrations are then used for DP3 policy training. After carefully tuning hyper-parameters such as horizons and color utilizations, we report the evaluation results in Tab.~\ref{tab:exp-dp3}, with the best completion rate of 40\%. We suspect that the limited network capacity of the 3D encoder of DP3 prevents it from modeling the diverse state-action pairs present in the real-world human teleoperated demonstrations, leading it to only handle a smaller set of state-action pairs under the axis-wise action representations. On the contrary, RISE can handle real-world demonstrations with various action representations and maintain satisfactory performances. Lastly, compared to the evaluation setup in the DP3 paper, we allow objects to be placed anywhere in the entire workspace. This results in a greater variation of object locations, making the task more challenging.

\subsection{Generalization Test}

We evaluate the generalization ability of different methods on the \textbf{\textit{Collect Pens}} task with 1 pen under different levels of environmental disturbances as follows.

\begin{itemize}
    \item[\textbf{L1}.] Generalize to objects with similar shapes but different colors. In this task, we replace the original green bowl with a pink one, which is denoted as \textbf{Bowl} in Tab.~\ref{tab:exp-general}.
    \item[\textbf{L2}.] Generalize to different light conditions in the environment, denoted as \textbf{Light} in Tab.~\ref{tab:exp-general}. 
    \item[\textbf{L3}.] Generalize to new workspace configurations in the environment. In this task, we elevate the workspace by 13cm to form a new workspace configuration, which is denoted as \textbf{Height} in Tab.~\ref{tab:exp-general}.
    \item[\textbf{L4}.] Generalize to new camera viewpoints, denoted as \textbf{CamView} in Tab.~\ref{tab:exp-general}.
\end{itemize}

The graphical illustration of the environmental disturbances and the evaluation results are shown in Tab.~\ref{tab:exp-general}. We can observe that the image-based policies can achieve a decent L1-level and some L2-level generalizations, but they cannot reach the L3-level and L4-level generalizations involving spatial transformations. Act3D, as a 3D policy, demonstrates good generalization up to L3-level disturbances; however, it nearly completely fails in the L4-level generalization test. RISE exhibits strong generalization abilities across all levels of testing, even in the most challenging L4-level tests involving changes in the camera view.

\section{Conclusion}
In this paper, we present RISE, an efficient end-to-end policy utilizing 3D perception for real-world robot manipulation. RISE compresses point clouds with a sparse 3D encoder, followed by sparse positional encoding and a transformer to obtain action features. The features are decoded into continuous actions by a diffusion head. RISE significantly outperforms currently representative 2D and 3D policies in multiple tasks, demonstrating great advantages in both accuracy and efficiency. Our ablations verify the effectiveness of 3D perception and the generalization of RISE under different levels of environmental disturbances. We hope our baseline inspires the integration of 3D perception into real-world policy learning.
\section{Acknowledgement}
This work was supported by the National Key Research and Development Project of China (No. 2022ZD0160102), the National Key Research and Development Project of China (No. 2021ZD0110704), Shanghai Artificial Intelligence Laboratory, XPLORER PRIZE grants.

\printbibliography

@inproceedings{zhao2023act,
	title={Learning Fine-Grained Bimanual Manipulation with Low-Cost Hardware},
	author={Zhao, Tony Z and Kumar, Vikash and Levine, Sergey and Finn, Chelsea},
	booktitle={Robotics: Science and Systems},
	year={2023}
}

@inproceedings{chi2023diffusion,
	title={Diffusion Policy: Visuomotor Policy Learning via Action Diffusion},
	author={Chi, Cheng and Feng, Siyuan and Du, Yilun and Xu, Zhenjia and Cousineau, Eric and Burchfiel, Benjamin and Song, Shuran},
	booktitle={Robotics: Science and Systems},
	year={2023}
}

@inproceedings{fang2023low,
  title={AirExo: Low-Cost Exoskeletons for Learning Whole-Arm Manipulation in the Wild},
  author={Fang, Hongjie and Fang, Hao-Shu and Wang, Yiming and Ren, Jieji and Chen, Jingjing and Zhang, Ruo and Wang, Weiming and Lu, Cewu},
  booktitle={IEEE International Conference on Robotics and Automation},
  year={2024}
}

@article{wang2024dexcap,
  title = {DexCap: Scalable and Portable Mocap Data Collection System for Dexterous Manipulation},
  author = {Wang, Chen and Shi, Haochen and Wang, Weizhuo and Zhang, Ruohan and Fei-Fei, Li and Liu, C. Karen},
  journal = {Proceedings of Robotics: Science and Systems (RSS)},
  year = {2024}
}

@inproceedings{rt1,
  author       = {Anthony Brohan and
                  Noah Brown and
                  Justice Carbajal and
                  Yevgen Chebotar and
                  Joseph Dabis and
                  Chelsea Finn and
                  Keerthana Gopalakrishnan and
                  Karol Hausman and
                  Alexander Herzog and
                  Jasmine Hsu and
                  Julian Ibarz and
                  Brian Ichter and
                  Alex Irpan and
                  Tomas Jackson and
                  Sally Jesmonth and
                  Nikhil J. Joshi and
                  Ryan Julian and
                  Dmitry Kalashnikov and
                  Yuheng Kuang and
                  Isabel Leal and
                  Kuang{-}Huei Lee and
                  Sergey Levine and
                  Yao Lu and
                  Utsav Malla and
                  Deeksha Manjunath and
                  Igor Mordatch and
                  Ofir Nachum and
                  Carolina Parada and
                  Jodilyn Peralta and
                  Emily Perez and
                  Karl Pertsch and
                  Jornell Quiambao and
                  Kanishka Rao and
                  Michael S. Ryoo and
                  Grecia Salazar and
                  Pannag R. Sanketi and
                  Kevin Sayed and
                  Jaspiar Singh and
                  Sumedh Sontakke and
                  Austin Stone and
                  Clayton Tan and
                  Huong Tran and
                  Vincent Vanhoucke and
                  Steve Vega and
                  Quan Vuong and
                  Fei Xia and
                  Ted Xiao and
                  Peng Xu and
                  Sichun Xu and
                  Tianhe Yu and
                  Brianna Zitkovich},
  title        = {{RT-1:} Robotics Transformer for Real-World Control at Scale},
  booktitle    = {Robotics: Science and Systems},
  year         = {2023}
}

@inproceedings{rt2,
  title = 	 {RT-2: Vision-Language-Action Models Transfer Web Knowledge to Robotic Control},
  author =       {Zitkovich, Brianna and Yu, Tianhe and Xu, Sichun and Xu, Peng and Xiao, Ted and Xia, Fei and Wu, Jialin and Wohlhart, Paul and Welker, Stefan and Wahid, Ayzaan and Vuong, Quan and Vanhoucke, Vincent and Tran, Huong and Soricut, Radu and Singh, Anikait and Singh, Jaspiar and Sermanet, Pierre and Sanketi, Pannag R. and Salazar, Grecia and Ryoo, Michael S. and Reymann, Krista and Rao, Kanishka and Pertsch, Karl and Mordatch, Igor and Michalewski, Henryk and Lu, Yao and Levine, Sergey and Lee, Lisa and Lee, Tsang-Wei Edward and Leal, Isabel and Kuang, Yuheng and Kalashnikov, Dmitry and Julian, Ryan and Joshi, Nikhil J. and Irpan, Alex and Ichter, Brian and Hsu, Jasmine and Herzog, Alexander and Hausman, Karol and Gopalakrishnan, Keerthana and Fu, Chuyuan and Florence, Pete and Finn, Chelsea and Dubey, Kumar Avinava and Driess, Danny and Ding, Tianli and Choromanski, Krzysztof Marcin and Chen, Xi and Chebotar, Yevgen and Carbajal, Justice and Brown, Noah and Brohan, Anthony and Arenas, Montserrat Gonzalez and Han, Kehang},
  booktitle = 	 {Conference on Robot Learning},
  pages = 	 {2165--2183},
  year = 	 {2023},
  organizations =    {PMLR}
}

@inproceedings{oxe,
  title={Open X-Embodiment: Robotic Learning Datasets and RT-X Models},
  author={Padalkar, Abhishek and Pooley, Acorn and Jain, Ajinkya and Bewley, Alex and Herzog, Alex and Irpan, Alex and Khazatsky, Alexander and Rai, Anant and Singh, Anikait and Brohan, Anthony and others},
  booktitle={IEEE International Conference on Robotics and Automation},
  year={2024}
}

@inproceedings{rh20t,
    title = {RH20T: A Comprehensive Robotic Dataset for Learning Diverse Skills in One-Shot},
    author = {Fang, Hao-Shu and Fang, Hongjie and Tang, Zhenyu and Liu, Jirong and Wang, Chenxi and Wang, Junbo and Zhu, Haoyi and Lu, Cewu},
    booktitle = {IEEE International Conference on Robotics and Automation},
    year = {2024}
}

@inproceedings{shridar2022peract,
  author    = {Mohit Shridhar and
               Lucas Manuelli and
               Dieter Fox},
  title     = {Perceiver-Actor: A Multi-Task Transformer for Robotic Manipulation},
  booktitle = {Conference on Robot Learning},
  pages     = {785--799},
  year      = {2022},
  organization={PMLR}
}

@inproceedings{zhu2023learning,
  title={Learning generalizable manipulation policies with object-centric 3d representations},
  author={Zhu, Yifeng and Jiang, Zhenyu and Stone, Peter and Zhu, Yuke},
  booktitle={7th Annual Conference on Robot Learning},
  year={2023}
}

@inproceedings{he2016resnet,
  title={Deep Residual Learning for Image Recognition},
  author={He, Kaiming and Zhang, Xiangyu and Ren, Shaoqing and Sun, Jian},
  booktitle={Proceedings of the IEEE Conference on Computer Vision and Pattern Recognition},
  pages={770--778},
  year={2016}
}

@inproceedings{choy2019minkowski,
  title={4D Spatio-Temporal ConvNets: Minkowski Convolutional Neural Networks},
  author={Choy, Christopher and Gwak, JunYoung and Savarese, Silvio},
  booktitle={Proceedings of the IEEE/CVF Conference on Computer Vision and Pattern Recognition},
  pages={3075--3084},
  year={2019}
}

@article{vaswani2017attention,
  title={Attention is All You Need},
  author={Vaswani, Ashish and Shazeer, Noam and Parmar, Niki and Uszkoreit, Jakob and Jones, Llion and Gomez, Aidan N and Kaiser, {\L}ukasz and Polosukhin, Illia},
  journal={Advances in Neural Information Processing Systems},
  volume={30},
  year={2017}
}

@inproceedings{rot6d,
  title={On the Continuity of Rotation Representations in Neural Networks},
  author={Zhou, Yi and Barnes, Connelly and Lu, Jingwan and Yang, Jimei and Li, Hao},
  booktitle={Proceedings of the IEEE/CVF Conference on Computer Vision and Pattern Recognition},
  pages={5745--5753},
  year={2019}
}

@inproceedings{gervet23act3d,
  author       = {Th{\'{e}}ophile Gervet and
                  Zhou Xian and
                  Nikolaos Gkanatsios and
                  Katerina Fragkiadaki},
  title        = {Act3D: 3D Feature Field Transformers for Multi-Task Robotic Manipulation},
  booktitle    = {Conference on Robot Learning},
  pages        = {3949--3965},
  organization    = {PMLR},
  year         = {2023}
}

@inproceedings{ze2023gnfactor,
  author       = {Yanjie Ze and
                  Ge Yan and
                  Yueh{-}Hua Wu and
                  Annabella Macaluso and
                  Yuying Ge and
                  Jianglong Ye and
                  Nicklas Hansen and
                  Li Erran Li and
                  Xiaolong Wang},
  title        = {GNFactor: Multi-Task Real Robot Learning with Generalizable Neural Feature Fields},
  booktitle    = {Conference on Robot Learning},
  pages        = {284--301},
  organization    = {PMLR},
  year         = {2023}
}

@inproceedings{guhur2022hiveformer,
  author       = {Pierre{-}Louis Guhur and
                  Shizhe Chen and
                  Ricardo Garcia Pinel and
                  Makarand Tapaswi and
                  Ivan Laptev and
                  Cordelia Schmid},
  title        = {Instruction-Driven History-Aware Policies for Robotic Manipulations},
  booktitle    = {Conference on Robot Learning},
  pages        = {175--187},
  organization    = {PMLR},
  year         = {2022}
}

@article{ho2020denoising,
  title={Denoising Diffusion Probabilistic Models},
  author={Ho, Jonathan and Jain, Ajay and Abbeel, Pieter},
  journal={Advances in Neural Information Processing Systems},
  volume={33},
  pages={6840--6851},
  year={2020}
}

@inproceedings{song21denoising,
  author       = {Jiaming Song and
                  Chenlin Meng and
                  Stefano Ermon},
  title        = {Denoising Diffusion Implicit Models},
  booktitle    = {The International Conference on Learning Representations},
  year         = {2021}
}

@inproceedings{janner2022planning,
  title={Planning with Diffusion for Flexible Behavior Synthesis},
  author={Janner, Michael and Du, Yilun and Tenenbaum, Joshua and Levine, Sergey},
  booktitle={International Conference on Machine Learning},
  pages={9902--9915},
  year={2022},
  organization={PMLR}
}

@inproceedings{jang2022bcz,
  author       = {Eric Jang and
                  Alex Irpan and
                  Mohi Khansari and
                  Daniel Kappler and
                  Frederik Ebert and
                  Corey Lynch and
                  Sergey Levine and
                  Chelsea Finn},
  title        = {{BC-Z:} Zero-Shot Task Generalization with Robotic Imitation Learning},
  booktitle    = {Conference on Robot Learning},
  pages        = {991--1002},
  publisher    = {PMLR},
  year         = {2021}
}

@inproceedings{xian2023chaineddiffuser,
  author       = {Zhou Xian and
                  Nikolaos Gkanatsios and
                  Th{\'{e}}ophile Gervet and
                  Tsung{-}Wei Ke and
                  Katerina Fragkiadaki},
  title        = {ChainedDiffuser: Unifying Trajectory Diffusion and Keypose Prediction for Robotic Manipulation},
  booktitle    = {Conference on Robot Learning},
  pages        = {2323--2339},
  organization = {PMLR},
  year         = {2023}
}

@inproceedings{pari2021surprising,
  author       = {Jyothish Pari and
                  Nur Muhammad (Mahi) Shafiullah and
                  Sridhar Pandian Arunachalam and
                  Lerrel Pinto},
  title        = {The Surprising Effectiveness of Representation Learning for Visual Imitation},
  booktitle    = {Robotics: Science and Systems},
  year         = {2022}
}

@inproceedings{vip,
  author       = {Yecheng Jason Ma and
                  Shagun Sodhani and
                  Dinesh Jayaraman and
                  Osbert Bastani and
                  Vikash Kumar and
                  Amy Zhang},
  title        = {{VIP:} Towards Universal Visual Reward and Representation via Value-Implicit Pre-Training},
  booktitle    = {International Conference on Learning Representations},
  year         = {2023}
}

@inproceedings{vc1,
  title={Where are We in the Search for an Artificial Visual Cortex for Embodied Intelligence?},
  author={Majumdar, Arjun and Yadav, Karmesh and Arnaud, Sergio and Ma, Yecheng Jason and Chen, Claire and Silwal, Sneha and Jain, Aryan and Berges, Vincent-Pierre and Abbeel, Pieter and Malik, Jitendra and others},
  booktitle={ICRA 2023 Workshop on Pretraining for Robotics},
  year={2023}
}

@inproceedings{mvp,
  author       = {Ilija Radosavovic and
                  Tete Xiao and
                  Stephen James and
                  Pieter Abbeel and
                  Jitendra Malik and
                  Trevor Darrell},
  title        = {Real-World Robot Learning with Masked Visual Pre-Training},
  booktitle    = {Conference on Robot Learning},
  pages        = {416--426},
  year         = {2022},
  organization={PMLR}
}

@inproceedings{r3m,
  author    = {Suraj Nair and
               Aravind Rajeswaran and
               Vikash Kumar and
               Chelsea Finn and
               Abhinav Gupta},
  title     = {{R3M:} {A} Universal Visual Representation for Robot Manipulation},
  booktitle = {Conference on Robot Learning},
  pages     = {892--909},
  year      = {2022},
  organization={PMLR}
}

@inproceedings{liv,
  author       = {Yecheng Jason Ma and
                  Vikash Kumar and
                  Amy Zhang and
                  Osbert Bastani and
                  Dinesh Jayaraman},
  title        = {{LIV:} Language-Image Representations and Rewards for Robotic Control},
  booktitle    = {International Conference on Machine Learning},
  pages        = {23301--23320},
  organization    = {PMLR},
  year         = {2023}
}

@inproceedings{goyal2023rvt,
  author       = {Ankit Goyal and
                  Jie Xu and
                  Yijie Guo and
                  Valts Blukis and
                  Yu{-}Wei Chao and
                  Dieter Fox},
  title        = {{RVT:} Robotic View Transformer for 3D Object Manipulation},
  booktitle    = {Conference on Robot Learning},
  pages        = {694--710},
  organization    = {PMLR},
  year         = {2023}
}

@inproceedings{pomerleau1988alvinn,
  title={ALVINN: An Autonomous Land Vehicle in a Neural Network},
  author={Pomerleau, Dean A},
  booktitle={Advances in Neural Information Processing Systems},
  volume={1},
  year={1988}
}

@article{gato,
  author       = {Scott E. Reed and
                  Konrad Zolna and
                  Emilio Parisotto and
                  Sergio G{\'{o}}mez Colmenarejo and
                  Alexander Novikov and
                  Gabriel Barth{-}Maron and
                  Mai Gimenez and
                  Yury Sulsky and
                  Jackie Kay and
                  Jost Tobias Springenberg and
                  Tom Eccles and
                  Jake Bruce and
                  Ali Razavi and
                  Ashley Edwards and
                  Nicolas Heess and
                  Yutian Chen and
                  Raia Hadsell and
                  Oriol Vinyals and
                  Mahyar Bordbar and
                  Nando de Freitas},
  title        = {A Generalist Agent},
  journal      = {Transactions on Machine Learning Research},
  year         = {2022}
}

@misc{octo,
  title={Octo: An Open-Source Generalist Robot Policy},
  author={Team, Octo Model and Ghosh, Dibya and Walke, Homer and Pertsch, Karl and Black, Kevin and Mees, Oier and Dasari, Sudeep and Hejna, Joey and Xu, Charles and Luo, Jianlan and others},
  year={2023}
}

@article{mees2022calvin,
  title={CALVIN: A Benchmark for Language-Conditioned Policy Learning for Long-Horizon Robot Manipulation Tasks},
  author={Mees, Oier and Hermann, Lukas and Rosete-Beas, Erick and Burgard, Wolfram},
  journal={IEEE Robotics and Automation Letters},
  volume={7},
  number={3},
  pages={7327--7334},
  year={2022},
  publisher={IEEE}
}

@article{james2020rlbench,
  title={RLBench: The Robot Learning Benchmark \& Learning Environment},
  author={James, Stephen and Ma, Zicong and Arrojo, David Rovick and Davison, Andrew J},
  journal={IEEE Robotics and Automation Letters},
  volume={5},
  number={2},
  pages={3019--3026},
  year={2020},
  publisher={IEEE}
}

@inproceedings{ha2023scaling,
  author       = {Huy Ha and
                  Pete Florence and
                  Shuran Song},
  title        = {Scaling Up and Distilling Down: Language-Guided Robot Skill Acquisition},
  booktitle    = {Conference on Robot Learning},
  pages        = {3766--3777},
  organization    = {PMLR},
  year         = {2023}
}

@inproceedings{mandlekar2021what,
  author    = {Ajay Mandlekar and
               Danfei Xu and
               Josiah Wong and
               Soroush Nasiriany and
               Chen Wang and
               Rohun Kulkarni and
               Li Fei{-}Fei and
               Silvio Savarese and
               Yuke Zhu and
               Roberto Mart{\'{\i}}n{-}Mart{\'{\i}}n},
  title     = {What Matters in Learning from Offline Human Demonstrations for Robot Manipulation},
  booktitle = {Conference on Robot Learning},
  pages     = {1678--1690},
  year      = {2021},
  organization={PMLR}
}

@inproceedings{mandlekar2018roboturk,
  title={RoboTurk: A Crowdsourcing Platform for Robotic Skill Learning through Imitation},
  author={Mandlekar, Ajay and Zhu, Yuke and Garg, Animesh and Booher, Jonathan and Spero, Max and Tung, Albert and Gao, Julian and Emmons, John and Gupta, Anchit and Orbay, Emre and others},
  booktitle={Conference on Robot Learning},
  pages={879--893},
  year={2018},
  organization={PMLR}
}

@inproceedings{james2022coarse,
  title={Coarse-to-Fine Q-Attention: Efficient Learning for Visual Robotic Manipulation via Discretisation},
  author={James, Stephen and Wada, Kentaro and Laidlow, Tristan and Davison, Andrew J},
  booktitle={Proceedings of the IEEE/CVF Conference on Computer Vision and Pattern Recognition},
  pages={13739--13748},
  year={2022}
}

@article{anygrasp,
  title={AnyGrasp: Robust and Efficient Grasp Perception in Spatial and Temporal Domains},
  author={Fang, Hao-Shu and Wang, Chenxi and Fang, Hongjie and Gou, Minghao and Liu, Jirong and Yan, Hengxu and Liu, Wenhai and Xie, Yichen and Lu, Cewu},
  journal={IEEE Transactions on Robotics},
  year={2023},
  publisher={IEEE}
}

@inproceedings{rahmatizadeh2018vision,
  title={Vision-Based Multi-Task Manipulation for Inexpensive Robots using End-to-End Learning from Demonstration},
  author={Rahmatizadeh, Rouhollah and Abolghasemi, Pooya and B{\"o}l{\"o}ni, Ladislau and Levine, Sergey},
  booktitle={IEEE International Conference on Robotics and Automation},
  pages={3758--3765},
  year={2018},
  organization={IEEE}
}

@inproceedings{wu20153d,
  title={3D ShapeNets: A Deep Representation for Volumetric Shapes},
  author={Wu, Zhirong and Song, Shuran and Khosla, Aditya and Yu, Fisher and Zhang, Linguang and Tang, Xiaoou and Xiao, Jianxiong},
  booktitle={Proceedings of the IEEE Conference on Computer Vision and Pattern Recognition},
  pages={1912--1920},
  year={2015}
}

@inproceedings{dai2017scannet,
  title={ScanNet: Richly-Annotated 3D Reconstructions of Indoor Scenes},
  author={Dai, Angela and Chang, Angel X and Savva, Manolis and Halber, Maciej and Funkhouser, Thomas and Nie{\ss}ner, Matthias},
  booktitle={Proceedings of the IEEE Conference on Computer Vision and Pattern Recognition},
  pages={5828--5839},
  year={2017}
}

@inproceedings{zhou2018voxelnet,
  title={VoxelNet: End-to-End Learning for Point Cloud Based 3D Object Detection},
  author={Zhou, Yin and Tuzel, Oncel},
  booktitle={Proceedings of the IEEE Conference on Computer Vision and Pattern Recognition},
  pages={4490--4499},
  year={2018}
}

@inproceedings{qi2017pointnet,
  title={PointNet: Deep Learning on Point Sets for 3D Classification and Segmentation},
  author={Qi, Charles R and Su, Hao and Mo, Kaichun and Guibas, Leonidas J},
  booktitle={Proceedings of the IEEE Conference on Computer Vision and Pattern Recognition},
  pages={652--660},
  year={2017}
}

@inproceedings{riegler2017octnet,
  title={OctNet: Learning Deep 3D Representations at High Resolutions},
  author={Riegler, Gernot and Osman Ulusoy, Ali and Geiger, Andreas},
  booktitle={Proceedings of the IEEE conference on computer vision and pattern recognition},
  pages={3577--3586},
  year={2017}
}

@article{wang2017cnn,
  title={O-CNN: Octree-Based Convolutional Neural Networks for 3D Shape Analysis},
  author={Wang, Peng-Shuai and Liu, Yang and Guo, Yu-Xiao and Sun, Chun-Yu and Tong, Xin},
  journal={ACM Transactions On Graphics},
  volume={36},
  number={4},
  pages={1--11},
  year={2017},
  publisher={ACM New York, NY, USA}
}

@inproceedings{zhao2021point,
  title={Point Transformer},
  author={Zhao, Hengshuang and Jiang, Li and Jia, Jiaya and Torr, Philip HS and Koltun, Vladlen},
  booktitle={Proceedings of the IEEE/CVF International Conference on Computer Vision},
  pages={16259--16268},
  year={2021}
}

@inproceedings{li2016vehicle,
  author       = {Bo Li and
                  Tianlei Zhang and
                  Tian Xia},
  title        = {Vehicle Detection from 3D Lidar Using Fully Convolutional Network},
  booktitle    = {Robotics: Science and Systems},
  year         = {2016}
}

@inproceedings{chen2017multi,
  title={Multi-View 3D Object Detection Network for Autonomous Driving},
  author={Chen, Xiaozhi and Ma, Huimin and Wan, Ji and Li, Bo and Xia, Tian},
  booktitle={Proceedings of the IEEE Conference on Computer Vision and Pattern Recognition},
  pages={1907--1915},
  year={2017}
}

@inproceedings{hamdi2021mvtn,
  title={MVTN: Multi-View Transformation Network for 3D Shape Recognition},
  author={Hamdi, Abdullah and Giancola, Silvio and Ghanem, Bernard},
  booktitle={Proceedings of the IEEE/CVF International Conference on Computer Vision},
  pages={1--11},
  year={2021}
}

@inproceedings{chen2023polarnet,
  author       = {Shizhe Chen and
                  Ricardo Garcia Pinel and
                  Cordelia Schmid and
                  Ivan Laptev},
  title        = {PolarNet: 3D Point Clouds for Language-Guided Robotic Manipulation},
  booktitle    = {Conference on Robot Learning},
  pages        = {1761--1781},
  organization    = {{PMLR}},
  year         = {2023}
}

@inproceedings{sgr23,
  author       = {Tong Zhang and
                  Yingdong Hu and
                  Hanchen Cui and
                  Hang Zhao and
                  Yang Gao},
  title        = {A Universal Semantic-Geometric Representation for Robotic Manipulation},
  booktitle    = {CoRL},
  series       = {Proceedings of Machine Learning Research},
  volume       = {229},
  pages        = {3342--3363},
  publisher    = {{PMLR}},
  year         = {2023}
}

@article{qian2022pointnext,
  title={PointNeXt: Revisiting PointNet++ with Improved Training and Scaling Strategies},
  author={Qian, Guocheng and Li, Yuchen and Peng, Houwen and Mai, Jinjie and Hammoud, Hasan and Elhoseiny, Mohamed and Ghanem, Bernard},
  journal={Advances in Neural Information Processing Systems},
  volume={35},
  pages={23192--23204},
  year={2022}
}

@inproceedings{pan20213d,
  title={3D Object Detection with PointFormer},
  author={Pan, Xuran and Xia, Zhuofan and Song, Shiji and Li, Li Erran and Huang, Gao},
  booktitle={Proceedings of the IEEE/CVF Conference on Computer Vision and Pattern Recognition},
  pages={7463--7472},
  year={2021}
}

@inproceedings{graham20183d,
  title={3D Semantic Segmentation with Submanifold Sparse Convolutional Networks},
  author={Graham, Benjamin and Engelcke, Martin and Van Der Maaten, Laurens},
  booktitle={Proceedings of the IEEE Conference on Computer Vision and Pattern Recognition},
  pages={9224--9232},
  year={2018}
}

@inproceedings{zhang2018deep,
  title={Deep Imitation Learning for Complex Manipulation Tasks from Virtual Reality Teleoperation},
  author={Zhang, Tianhao and McCarthy, Zoe and Jow, Owen and Lee, Dennis and Chen, Xi and Goldberg, Ken and Abbeel, Pieter},
  booktitle={IEEE International Conference on Robotics and Automation},
  pages={5628--5635},
  year={2018},
  organization={IEEE}
}

@inproceedings{sundermeyer2021contact,
  title={Contact-Graspnet: Efficient 6-DoF Grasp Generation in Cluttered Scenes},
  author={Sundermeyer, Martin and Mousavian, Arsalan and Triebel, Rudolph and Fox, Dieter},
  booktitle={IEEE International Conference on Robotics and Automation},
  pages={13438--13444},
  year={2021},
  organization={IEEE}
}

@article{sohn2015learning,
  title={Learning Structured Output Representation using Deep Conditional Generative Models},
  author={Sohn, Kihyuk and Lee, Honglak and Yan, Xinchen},
  journal={Advances in Neural Information Processing Systems},
  volume={28},
  year={2015}
}

@article{mildenhall2021nerf,
  title={NeRF: Representing Scenes as Neural Radiance Fields for View Synthesis},
  author={Mildenhall, Ben and Srinivasan, Pratul P and Tancik, Matthew and Barron, Jonathan T and Ramamoorthi, Ravi and Ng, Ren},
  journal={Communications of the ACM},
  volume={65},
  number={1},
  pages={99--106},
  year={2021},
  publisher={ACM New York, NY, USA}
}

@article{ze2024dp3,
    title={3D Diffusion Policy: Generalizable Visuomotor Policy Learning via Simple 3D Representations},
    author={Yanjie Ze and Gu Zhang and Kangning Zhang and Chenyuan Hu and Muhan Wang and Huazhe Xu},
    journal={Proceedings of Robotics: Science and Systems (RSS)},
    year={2024}
}

@inproceedings{shen2023distilled,
  author       = {William Shen and
                  Ge Yang and
                  Alan Yu and
                  Jansen Wong and
                  Leslie Pack Kaelbling and
                  Phillip Isola},
  title        = {Distilled Feature Fields Enable Few-Shot Language-Guided Manipulation},
  booktitle    = {Conference on Robot Learning},
  pages        = {405--424},
  organization    = {{PMLR}},
  year         = {2023}
}

@inproceedings{ye2023featurenerf,
  title={FeatureNeRF: Learning Generalizable NeRFs by Distilling Foundation Models},
  author={Ye, Jianglong and Wang, Naiyan and Wang, Xiaolong},
  booktitle={Proceedings of the IEEE/CVF International Conference on Computer Vision},
  pages={8962--8973},
  year={2023}
}

@article{driess2022reinforcement,
  title={Reinforcement Learning with Neural Radiance Fields},
  author={Driess, Danny and Schubert, Ingmar and Florence, Pete and Li, Yunzhu and Toussaint, Marc},
  journal={Advances in Neural Information Processing Systems},
  volume={35},
  pages={16931--16945},
  year={2022}
}

% not for IROS, but can be included in preprint version
\clearpage
\section*{APPENDIX}
\subsection{Tasks Parameters}

We list the parameters of the demonstrations for different tasks in this paper in Tab.~\ref{tab:task-params}. The axis-wise action representation is implemented with keyboard teleoperation (one key to control movement, or rotation, or gripper action in each direction). We observe that although the axis-wise action representation results in fewer steps during demonstrations, its teleoperation time was approximately 3x as long as that of the natural teleoperation, aligning with the findings in~\cite{mandlekar2018roboturk}.

\begin{table}[htbp]
\centering
        \footnotesize
        \setlength\tabcolsep{2pt}
        \renewcommand{\arraystretch}{1.2}
        \begin{tabular}{ccccc} \hline
        \textbf{Task Name} & \textbf{Notes} & \textbf{\# Demos} & \textbf{Avg. Steps} & \begin{tabular}{c} \textbf{Avg. Teleop.}\\ \textbf{Time (s)}\end{tabular} \\ \hline
        \multirow{7}{*}{\textbf{\textit{Collect Cups}}} & 1 cup & 10 & 117.4 & 19.37\\
        & 2 cups & 10 & 225.0 & 34.73 \\
        & 3 cups & 10 & 345.3 & 54.84 \\
        & 4 cups & 10 & 451.4 & 71.07 \\
        & 5 cups & 10 & 520.0 & 76.02  \\ \cline{2-5}
        & \begin{tabular}{c} $^*$ 1 cup, \\ natural action \end{tabular} & 50 & 102.7 & 17.06 \\
        & \begin{tabular}{c} $^*$ 1 cup, \\ axis-wise action \end{tabular} & 50 & 30.2 & 45.93 \\
        \hline
        \multirow{5}{*}{\textbf{\textit{Collect Pens}}} & 1 pen & 10 & 179.4 & 52.47 \\
        & 2 pens & 10 & 278.2 & 62.71 \\
        & 3 pens & 10 & 411.5 & 91.88 \\
        & 4 pens & 10 & 556.1 & 124.22 \\
        & 5 pens & 10 & 694.1 & 157.15 \\
        \hline
        \textbf{\textit{Pour Balls}} & & 50 & 185.4 & 50.69 \\ \hline
        \textbf{\textit{Push Block}} & & 50 & 204.3 & 51.72 \\ \hline
        \textbf{\textit{Push Ball}} & & 50 & 223.1 & 46.00 \\ \hline
        \multirow{3}{*}{\textbf{\textit{Stack Blocks}}} & 2 blocks & 10 & 148.6 & 32.06 \\ 
        & 3 blocks & 20 & 286.2 & 59.22 \\
        & 4 blocks & 20 & 401.8 & 79.83 \\
        \hline
        \end{tabular}
    \caption{Parameters of the collected demonstrations for different tasks. ``Avg. Teleop. Time'' stands for the average teleoperation time for collecting one demonstration. $^*$ denotes that these data are only used for the comparison experiments with DP3.}\label{tab:task-params}\vspace{-0.1cm}
\end{table}

The evaluation settings for different tasks is summarized in Tab.~\ref{tab:exp-params}. Compared to the average steps in demonstrations, the maximum steps in evaluations prove to be sufficient.

\begin{table}[htbp]
\centering
        \footnotesize
        \setlength\tabcolsep{2pt}
        \renewcommand{\arraystretch}{1.2}
        \begin{tabular}{ccccc} \hline
        \textbf{Task Name} & \textbf{Notes} & \textbf{\# Trials} & \textbf{Max. Steps} & \textbf{Max. Keyframes} \\ \hline
        \multirow{5}{*}{\textbf{\textit{Collect Cups}}} & 1 cup & 10 & 300 & 20 \\
        & 2 cups & 10 & 600 & 40 \\
        & 3 cups & 10 & 900 & 60 \\
        & 4 cups & 10 & 1200 & 80 \\
        & 5 cups & 10 & 1500 & 100 \\  \hline
        \multirow{5}{*}{\textbf{\textit{Collect Pens}}} & 1 pen & 10 & 300 & 20 \\
        & 2 pens & 10 & 600 & 40 \\
        & 3 pens & 10 & 900 & 60 \\
        & 4 pens & 10 & 1200 & 80 \\
        & 5 pens & 10 & 1500 & 100 \\ \hline
        \textbf{\textit{Pour Balls}} & & 20 & 1200 & N/A \\ \hline
        \textbf{\textit{Push Block}} & & 20 & 1200 & N/A \\ \hline
        \textbf{\textit{Push Ball}} & & 20 & 1200 & N/A \\ \hline
        \multirow{3}{*}{\textbf{\textit{Stack Blocks}}} & 2 blocks & 10 & 600 & N/A \\ 
        & 3 blocks & 10 & 1200 & N/A \\
        & 4 blocks & 10 & 1800 & N/A \\
        \hline
        \end{tabular}
    \caption{Evaluation settings for different tasks.}\label{tab:exp-params}\vspace{-0.2cm}
\end{table}

\subsection{Implementation Details}
\textbf{Data Processing.} The point cloud is created from a single-view RGB-D image. Both input point clouds and output actions are in the camera coordinate system. We crop the point clouds with the range of $x,y\in [-0.5\mathrm{m},0.5\mathrm{m}]$, $z\in[0\mathrm{m},1\mathrm{m}]$, and normalize the translation values to $[-1,1]$ with the range of $x,y\in [-0.35\mathrm{m},0.35\mathrm{m}]$, $z\in[0\mathrm{m},0.7\mathrm{m}]$. The gripper width is normalized to $[-1,1]$ using the range of $[0\mathrm{m},0.11\mathrm{m}]$.

\textbf{Network.} The sparse 3D encoder is implemented based on MinkowskiEngine~\cite{choy2019minkowski} with a voxel size of 5mm, which outputs a set of point feature vectors at the dimension of 512. For sparse positional encoding, we set $v=5\mathrm{mm}$ and $c=400$. The transformer contains 4 encoding blocks and 1 decoding block, with $d_\text{model}=512$ and $d_\text{ff}=2048$. The dimension of the readout token is 512. We employ a CNN-based diffusion head \cite{chi2023diffusion} with 100 denoising iterations for training and 20 iterations for inference. The output action horizon is 20.

\textbf{Training.} RISE is trained on 2 Nvidia A100 GPUs with a batch size of 240, an initial learning rate of 3e-4, and a warmup step of 2000. The learning rate is decayed by a cosine scheduler. During training, the point clouds are randomly translated by $[-0.2\mathrm{m},0.2\mathrm{m}]$ along X/Y/Z-axis, and randomly rotated by $[-30^{\circ},30^{\circ}]$ around X/Y/Z-axis.

\textbf{Baseline.} ACT~\cite{zhao2023act}, Diffusion Policy~\cite{chi2023diffusion}, Act3D~\cite{gervet23act3d} and DP3~\cite{ze2024dp3} are trained based on the official implementations. The Diffusion Policy baseline takes ResNet18 as the visual encoder and adopts a CNN-based backbone. For the Act3D baseline, we implement a simple planner for \textit{pick-and-place} tasks to avoid collisions, which decouples an action into a horizontal one and a vertical one. It follows a heuristic rule: the horizontal action precedes the downward one while it follows the upward one. For ACT (3D), we replace the image tokens with the point tokens. For Diffusion Policy (3D), we employ an AvgPooling layer to get the observation embedding from point features.

\subsection{Discussions about Action Representations}

\begin{figure*}
    \centering
    \includegraphics[width=\linewidth]{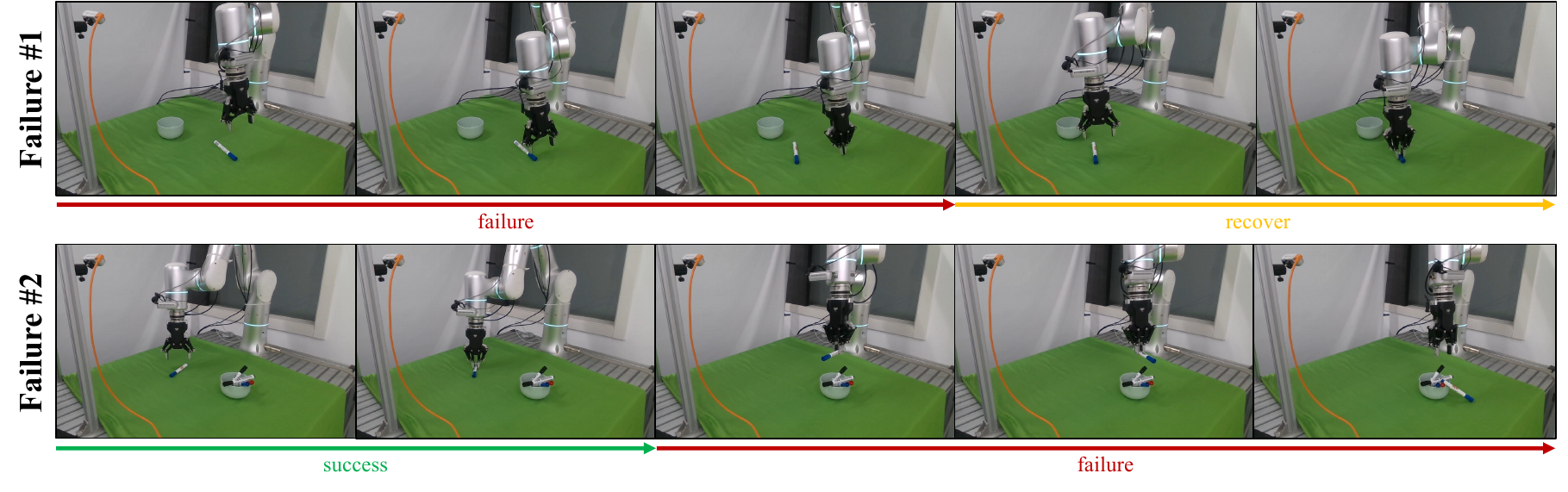}
    \vspace{-0.6cm}
    \caption{Failure cases of the \textbf{\textit{Collect Pens}} task in the experiments. }
    \label{fig:failure}
    \vspace{-0.4cm}
\end{figure*}

\textbf{Axis-wise.} (Tab.~\ref{tab:exp-dp3} (blue)) Axis-wise action representation assumes that only one axis-wise movement is conducted in each step (typically one of the translations along the X/Y/Z-axis, the rotation around the X/Y/Z-axis, and the gripper opening/closing). Demonstrations with axis-wise action representations are usually collected via teleoperation with low frequency, like keyboard teleoperations. 

\textbf{Natural.} (Tab.~\ref{tab:exp-dp3} (red)) Natural action representation allows composite movement patterns in each step (that is, the robot can simultaneously translate, rotate, and open or close the gripper in one step). Demonstrations with natural action representations are usually collected via teleoperations with high frequency, like teleoperations with haptic devices. 

\textbf{Discussions.} Due to only one non-zero value for each action at any step, axis-wise action representations are easy to learn. However, this ease of learning can introduce noticeable induction biases in the learned policy, resulting in a lack of action diversity. Moreover, the axis-wise action representation increases the difficulty of representing complex trajectories, resulting in the limited generalization capability of the learned policy. On the contrary, the natural action representation is more challenging to learn than the axis-wise action representation, the learned policy can exhibit more natural action trajectories. Furthermore, the natural action representation aligns more closely with the patterns of human action execution, thus adopting natural actions can enhance data collection efficiency, as illustrated in Tab.~\ref{tab:task-params}. Therefore, we adopt natural action representation in our collected real-world demonstrations.

\subsection{Failure Cases and Recovery} 

In this section, we take the \textbf{\textit{Collect Pens}} task as an example and illustrate the failure cases of RISE during experiments in Fig.~\ref{fig:failure}. We observe that failure cases are mainly caused by inaccurate positions during picking (Failure \#1) and placing (Failure \#2). We found that RISE can automatically correct some failure scenarios, such as instances where the pen is inadvertently moved due to imprecise positioning during grasping (Failure \#1). In contrast, many keyframe-based methods~\cite{gervet23act3d, shridar2022peract, xian2023chaineddiffuser} lack the ability to offer immediate recovery actions for failures, potentially leading to the exacerbation of errors.

\end{document}